\documentclass[twoside,11pt]{article}

\usepackage{jmlr2e}
\usepackage{soul}
\usepackage{graphicx}
\usepackage{amsmath}
\usepackage{booktabs}
\usepackage{algorithm}
\usepackage{algorithmic}
\usepackage{fancybox}

\usepackage{amsfonts}
\usepackage{subfigure}
\usepackage{verbatim}
\usepackage{amssymb}
\usepackage{verbatim}
\usepackage{enumitem}
\usepackage{sidecap,color}
\usepackage[square,numbers]{natbib} 

\def \textM {{\text{M}}}
\def \textS {{\text{S}}}

\def\true{{\text{true}}}
\def\EX{{\mathbb{E}}}

\def\bw{\mathbf{w}}
\def\mbI{\mathbb{I}}

\def \S {\mathbf{S}}

\def \R {\mathbb{R}}

\def \w {\mathbf{w}}

\def \x {\mathbf{x}}

\def \E {\mathrm{E}}

\def \x {\mathbf{x}}

\def \1 {\mathbf{1}}

\def \z {\mathbf{z}}

\def \y {\mathbf{y}}

\def \I {\mathbb{I}}

\def \y {\mathbf{y}}
\def \E {\mathrm{E}}
\def \x {\mathbf{x}}

\def \z {\mathbf{z}}

\def \w {\mathbf{w}}

\def \R {\mathbb{R}}
\def \S {\mathcal{S}}

\def \I {\mathbb{I}}

\usepackage{amssymb}
\usepackage{pifont}

\newlength\myindent
\jmlrheading{~}{~}{~}{~}{~}{~}{~}

\ShortHeadings{~}{~}
\firstpageno{1} 
\date{\today}

\begin{document}

\title{Large-scale Robust Deep AUC Maximization: A New Surrogate Loss and Empirical Studies on Medical Image Classification}

\author{\name Zhuoning Yuan$^\dagger$
        \email zhuoning-yuan@uiowa.edu \\
        \name Yan Yan$^\ddagger$ 
        \email yanyan.tju@gmail.com\\\
        \name Milan Sonka$^\dagger$
        \email milan-sonka@uiowa.edu\\
        \name Tianbao Yang$^\dagger$
        \email tianbao-yang@uiowa.edu  \\
        \addr $\dagger$Department of Computer Science, The University of Iowa, IA 52242 \\
        \addr $\ddagger$ School of Electrical Engineering\& Computer Science, Washington State University, WA 99163 \\
        }

\maketitle
\vspace{-0.6in}
\vspace{0.05in}

\begin{abstract}
\underline{D}eep \underline{A}UC \underline{M}aximization (DAM) is a new paradigm for learning a deep neural network  by maximizing the AUC score  of the model on a dataset. Most previous works of AUC maximization focus on the perspective of optimization by designing efficient stochastic algorithms, and studies on generalization performance of large-scale DAM on difficult tasks are missing. In this work, we aim to make  DAM more practical for interesting real-world applications (e.g., medical image classification). First, we propose a new {\bf margin-based min-max surrogate loss} function for the AUC score (named as the AUC min-max-margin loss or simply AUC margin loss for short). It is {\bf more robust} than the commonly used AUC square loss, while enjoying the same advantage in terms of large-scale stochastic optimization.  Second, we conduct extensive empirical studies of our DAM method on four difficult medical image classification tasks, namely (i) classification of chest x-ray images for identifying many threatening diseases, (ii)  classification of images of skin lesions for identifying melanoma, {(iii) classification of mammogram for breast cancer screening, and  (iv) classification of microscopic images for identifying tumor tissue}.  Our studies demonstrate that the proposed DAM method improves the performance of optimizing cross-entropy loss by a large margin, and also achieves better performance than optimizing the existing AUC square loss on these medical image classification tasks.   Specifically, our DAM method has achieved {\bf the 1st place} on Stanford {\bf CheXpert} competition on Aug. 31, 2020.    To the best of our knowledge, this is the first work that makes DAM succeed  on large-scale medical image datasets. We also conduct extensive ablation studies to demonstrate the advantages of the new AUC margin loss over the AUC square loss on benchmark datasets. The proposed method is implemented in our open-sourced library LibAUC (\url{www.libauc.org}) whose github address is~\url{https://github.com/Optimization-AI/LibAUC}.  

\end{abstract}

\begin{figure}[t!]
\centering
\includegraphics[width=0.6\textwidth]{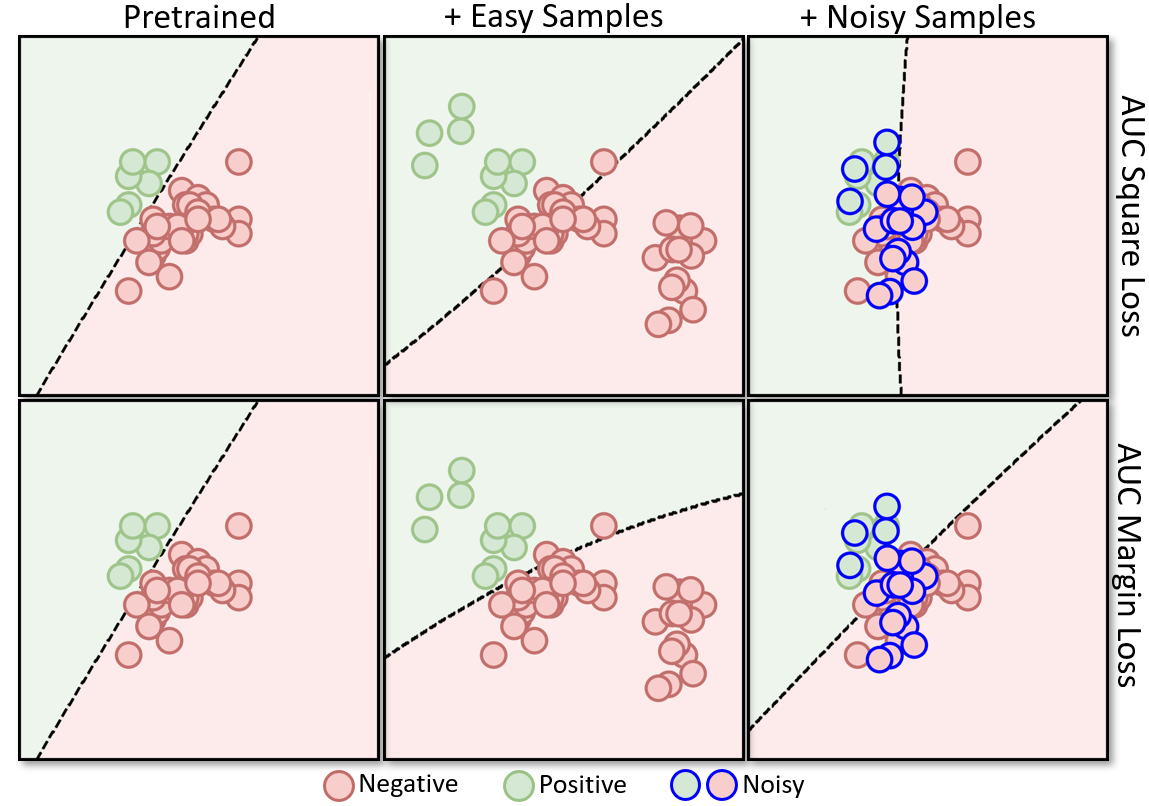}
\caption{An illustrative example for optimizing different AUC losses on a toy data for learning a two-layer neural network with ELU activation. The top row is optimizing the AUC square loss and the bottom row is optimizing the new AUC margin loss. The first column depicts the {initial} decision boundary (dashed line) pre-trained on a set of examples.  In the middle column, we add some easy examples to the training set and retrain the model by optimizing the AUC loss. In the last column, we add some noisily labeled data (blue circled data) to the training set and retrain the model by optimizing the AUC loss. The results demonstrate the new AUC margin loss is more robust than the AUC square loss. }
\label{fig:square_loss_drawback}
\vspace*{-0.2in}
\end{figure}

\section{Introduction}

\vspace*{-0.1in}In the last decade, we have seen  great progress in deep learning (DL) techniques for medical image classification driven by {\bf large-scale  medical datasets}.  For example, Stanford machine learning  group led by  Andrew Ng has collected and released a high-quality  large-scale Chest X-Ray dataset for detecting chest and lung diseases, which contains 224,316 high-quality X-rays images from 65,240 patients~\cite{irvin2019chexpert}. 
Various deep learning methods have been designed and evaluated on this dataset  by participating the {\bf CheXpert} competition organized by Stanford ML group~\cite{irvin2019chexpert}, and many of them have achieved radiologist-level performance on detecting certain related diseases. Esteva et al.~\cite{esteva2017dermatologist} have trained a CNN using a dataset of 129,450 clinical images consisting of 2,032 different diseases, and  achieved dermatologist-level performance for classification of skin lesions. Wu et al.~\cite{wu2019deep} have trained a deep neural network for breast cancer screening on a  large-scale medical dataset, which  includes 229,426 digital screening mammography exams (1,001,093 images) from 141,473 patients. Their model is as accurate as an experienced radiologist. Despite these great efforts, an important question remains: \\
\shadowbox{\begin{minipage}[t]{0.95\columnwidth} {\it ``Can we design a generic  method that can further improve the performance of DL on these medical datasets without relying on domain knowledge''}? \end{minipage}}

In this paper, we provide an affirmative answer to this question. Our solution is to optimize a novel loss for DL instead of optimizing the standard cross-entropy loss in the previous works. In particular, we choose to maximize the AUC score (a.k.a  {\bf the area under the ROC curve}) for DL. There are several benefits of maximizing AUC score  over minimizing the cross-entropy loss. First, in medical classification tasks the AUC score is the default metric for evaluating and comparing different methods. Directly maximizing AUC score can potentially lead to the largest improvement in the model's performance. Second, the datasets in medical image classification tasks are usually imbalanced (e.g., the number of malignant cases is usually much less than benign cases).  AUC  is more suitable for handling imbalanced data distribution since maximizing AUC aims to rank the predication score of any positive data higher than any negative data. However, AUC maximization is much more challenging than minimizing mis-classifcation error since AUC is much more sensitive to model change.  A simple example in Appendix \ref{section:example_sensitivity_auc_accuracy} shows that by changing the prediction scores of a few examples, the mis-classification error rate keep unchanged but the AUC score drops significantly.   

AUC maximization has been studied in the community of machine learning~\cite{gao2015consistency, ying2016stochastic, liu2019stochastic,joachims2005support, gao2013one}. However, existing methods for AUC maximization are still not satisfactory for practical use. The foremost  challenge for AUC maximization is to determine a surrogate loss for the AUC score. A naive way is to use a pairwise surrogate loss based on the definition of the AUC score. However, optimizing a generic pairwise loss on training data suffers from a severe scalability issue, which makes it not practical for DL on large-scale datasets. Several studies have made attempts to address the scalability issue~\cite{joachims2005support,Zhaoicml11,ying2016stochastic, liu2019stochastic}. One promising solution is to  maximize the pairwise square loss for AUC by utilizing its special form~\cite{ying2016stochastic,liu2019stochastic}. However, our study reveals that the AUC square loss  has adverse effect when trained with easy data and is sensitive to the noisy data.   

To address these issues, we propose a new margin-based surrogate loss in the min-max form for AUC (referred to as the AUC min-max-margin loss and the AUC margin loss for short), which is inspired by addressing the two issues of the AUC square loss. In particular, the AUC margin loss has two features that can alleviate the two issues, making it more robust to  noisy data and not  adversely affected by easy data. We will explain it with more details in the technical section and use a toy example in Figure 1 to illustrate the robustness of AUC margin loss over AUC square loss. Moreover, the min-max form of the AUC margin loss make it enjoy the same benefit as the AUC square loss in terms of scalability, making it more attractive than conventional pairwise margin-based surrogate loss for AUC maximization. In particular, we are able to directly employ  existing large-scale optimization algorithms~\cite{guo2020fast} designed for maximizing the AUC square loss to maximize our AUC margin loss with one line change of the code. 

To demonstrate the effectiveness of our deep AUC maximization method, we conduct empirical studies on {four difficult medical image classification tasks, namely classification of X-ray images for detecting chest diseases, classification of images of skin lesions, classification of mammograms for breast cancer screening and classification of microscopic images of tumor tissue}. Our deep AUC maximization method has achieved great success on these difficult tasks. Specifically, we achieved  {\bf the 1st place} on Stanford {\bf CheXpert} competition on Aug. 31, 2020, and \textbf{Top 1\%} rank on  Kaggle 2020 {\bf Melanoma} classification competition. In CheXpert competition, our method is ranked 1 out of 150+ submissions, with a 2\%+ improvement over Stanford baseline on a private testing data. 
In Kaggle competition, our ensembled model is ranked 33 out of 3314 teams. However, our best single model is better than the winning team's best model by more than 2\%. 
Besides these medical  tasks, we also conduct extensive ablation studies on benchmark datasets to compare the proposed AUC margin loss with the AUC square loss and traditional classification losses including cross-entropy and focal loss. Before ending this section, we summarize {\bf our contributions} below: 

\begin{itemize}[leftmargin=*]
\vspace*{-0.1in}
\item  We proposed a new robust surrogate loss for AUC maximization, which is more robust than the AUC square loss but enjoys the same benefit of large-scale optimization. 
\vspace*{-0.1in}
\item We conducted extensive empirical studies of the DAM method on a broad range of medical image classification data, and demonstrated its  superb performance compared with standard DL methods.
\end{itemize}
\vspace*{-0.1in}To the best of our knowledge, this is the first comprehensive study of DAM on large-scale medical image classification datasets. 

\section{Related Work}
\noindent {\bf Optimizing Pairwise Surrogate loss.} Based on the definition of AUC, many studies consider to optimize a pairwise surrogate loss for AUC~\cite{gao2015consistency, ying2016stochastic, liu2019stochastic}. Joachims et al~\cite{joachims2005support} proposed a SVM method for optimizing the AUC measure, which has a complexity of $O(n^2)$ for a dataset with $n$ examples. Many later studies tried to improve the efficiency of optimizing a pairwise surrogate loss of AUC. Herschtal et al.~\cite{herschtal2004optimising} proposed an approximate objective for empirical pairwise loss of AUC by using  partial pairs. In particular, for each negative data they only constructed  a pairwise loss with only one positive data. However, the quality of  such approximation highly depends on the properties of the dataset. When the examples have large intra-variance, their objective could yield poor performance. Zhao et al.~\cite{Zhaoicml11} proposed an online method for AUC maximization by maintaining a data buffer for storing some historical positive and negative data, and constructed an approximate AUC score by pairing a newly received data with all data in the buffer. However, analysis shows that such data buffer needs to be very large in order to make the algorithm has a small regret.

\vspace*{-0.01in}
 \noindent{\bf Optimizing Pairwise Square loss}.  Pairwise square loss is an exception, which has a unique property to enable one to design efficient stochastic algorithms for large-scale data~\cite{gao2013one,ying2016stochastic,liu2018fast, natole2018stochastic}. 
 In particular, Ying et al.~\cite{ying2016stochastic} formulated the minimization of the pairwise square loss into an equivalent min-max optimization problem, which allows them to develop efficient stochastic algorithms without explicitly constructing and handling pairs of positive and negative data. Several papers tried to improve the convergence rate for solving  the min-max optimization problems~\cite{liu2018fast, natole2018stochastic}. 
 
\vspace*{-0.01in}
\noindent{\bf Deep AUC Maximization (DAM)}.  Most of the studies mentioned above are for learning a linear model. Recently, there are some emerging studies on DAM.  In~\cite{sulam2017maximizing}, the authors considered DAM for learning a deep neural network based on an online buffered gradient method proposed by~\cite{Zhaoicml11}, and applied it to classification of breast cancer based on imbalanced mammogram images. Nevertheless, the issue of this approach is that it cannot scale to large datasets as it requires a large buffer to store positive and negative samples at each iteration for computing an approximate AUC score. Hence, they only consider datasets with few thousand medical images.  Recently, \cite{liu2019stochastic, guo2020fast}  proposed efficient stochastic non-convex min-max optimization algorithms for DAM by solving the corresponding min-max objective of the AUC square loss.  Their algorithms can scale up to  hundreds of thousands of training examples.  \cite{DBLP:journals/corr/abs-2005-02426,DBLP:conf/icml/YuanGXYY21} proposed  federated learning algorithms for distributed DAM.  
However, all of these studies have neglected the deficiencies  of the square loss for AUC maximization. To the best of our knowledge, this is the first work that analyzes the deficiencies of AUC square loss and proposes a better solution. 

\setlength{\abovedisplayskip}{0pt}
\setlength{\belowdisplayskip}{1pt}

\section{Method}
\noindent{\bf Notations.} Let $\mathbb I(\cdot)$ be an indicator function of a predicate, $[s]_+ = \max(s, 0)$.  Let $\S =\{(\x_1, y_1), \ldots, (\x_n, y_n)\}$ denote a set of training data, where $\x_i$ represents an input training example (e.g., an image), and $\y_i\in\{1, -1\}$ denotes its corresponding label (e.g., the indicator of a certain disease). For notational simplicity, we use $\z=(\x, y)$.  Let $\w\in \R^d$ denote the parameters of the deep neural network to be learned, and let $h_\w(\x) = h(\w, \x)$ denote the prediction of the neural network on an input data $\x$. The standard approach of deep learning is to define a loss function on individual data by $L(\w; \x, y) = \ell(h_\w(\x), y)$, where $\ell(\hat y, y)$ is a surrogate loss function of the misclassification error (e.g., cross-entropy loss), and to minimize the empirical loss $\min_{\w\in\R^d}\frac{1}{n}\sum_{i=1}^nL(\w; \x_i, y_i)$. However, this standard  approach is easily misled by the imbalanced distribution of training images in medical datasets. In medical applications, a more favorable  metric for comparing and evaluating different classifiers is  {\bf AUC}. It has been shown that the algorithms designed to minimize the misclassification error rate may not lead to maximization of AUC~\cite{NIPS2003_2518}. 

\subsection{Background on Scalable AUC Maximization}

Existing works of AUC maximization consider the following definition of AUC that is equivalent to the Wilcoxon-Mann-Whitney statistic \cite{hanley1982meaning, clemenccon2008ranking}:
\begin{align}\label{eqn:zeroone} 
\hspace*{-0.1in}\text{AUC}(\bw) & =   \Pr(h_\bw (\x) \ge  h_\bw(\x')| y =1, y'=-1 ) \\
&= \EX\bigl[\mbI(h_\bw (\x) - h_\bw(\x')\ge 0) \big | y=1, y'=-1\bigr].\nonumber
\end{align} 
It is interpreted that the  AUC score is the probability of a positive sample   ranking higher than a negative sample. 

For optimization purpose,  the indicator function in the above definition of AUC is usually replaced by a {\em convex surrogate loss} $\ell: \R \to \R^+$ which satisfies $\mbI(h_\bw (\x) - h_\bw(\x')< 0) \le \ell(h_\bw (\x) - h_\bw(\x')).$ 
As a result,  many existing works formulate the AUC maximization on a training data $\S$ as

\begin{equation}\label{eq:auc-emp}
\min_{\bw\in \R^d} \frac{1}{N_+N_-} \sum_{\x\in \S_+}\sum_{\x'\in \S_-}\ell(h_\bw (\x) - h_\bw(\x')),
\end{equation} 
where  $\S_+, \S_-$ denote the set of positive and negative examples, and $N_+, N_-$ denote their size, respectively. 
Nonetheless, directly optimizing the above formulation is not scalable to large datasets as the complexity could be as worse as $O(n^2)$ due to there are $O(n^2)$ pairs, where $n$ is the total number of examples. 

To address the scalability issue, existing studies have proposed some promising solutions. One solution that attracts great attention is to optimize the square loss due to its algorithmic simplicity.   With a square loss  $\ell(h_\bw (\x) - h_\bw(\x'))= (1 -h_\bw (\x) + h_\bw(\x') )^2$ as the surrogate loss of AUC,  it was shown that  the objective  is equivalent to  the following  min-max problem~\cite{ying2016stochastic}:  
\begin{align}\label{opt:spp}
\min_{\w\in\R^d \atop (a,b)\in\R^2}\max_{\alpha\in\R}f\left(\w,a,b,\alpha\right):=\EX_{\z}\left[F\left(\w,a,b,\alpha;\z\right)\right],
\end{align}
where $\z=(\x,y)\in\S$ is a random sample, and 
\begin{align}\label{eqn:AUCF}	&F(\w,a,b,\alpha;\z)=(1-p)\left(h_\w(\x)-a\right)^2\mathbb{I}_{[y=1]}\\
&+p(h_\w(\x)-b)^2\mathbb{I}_{[y=-1]}-p(1-p)\alpha^2\notag\\
&+2\alpha\left( p(1-p) + p h_\w(\x)\mathbb{I}_{[y=-1]}-(1-p)h_\w(\x)\mathbb{I}_{[y=1]}\right)\notag, 
\end{align}
and $p = \Pr(y=1)$.  
Since the objective function in the above formulation is decomposable over individual examples, hence it enables  one to develop efficient primal-dual stochastic algorithms for updating the model parameter $\w$ without explicitly constructing positive-negative pairs.  Several studies have developed efficient stochastic algorithms for solving the above min-max formulation, which are able to scale to hundreds of thousands of examples \cite{ying2016stochastic,liu2018fast,liu2019stochastic}.  

\subsection{Drawbacks of the AUC Square Loss}
\label{section:drawbacks_of_AUC_square_loss}
\vspace{-0.05in}
Although the AUC square loss makes AUC maximization scalable, it has two issues that have been ignored by existing studies. In particular, it has adverse effect when trained with well-classified data (i.e., easy data), and is sensitive to noisily labeled data (i.e., noisy data). Below, we will elaborate these two issues by considering a linear model $h_\w(\x) = \w^{\top}\x$ for illustration and understand these issues from the viewpoint of stochastic gradient update. We give a one-dimensional data in Appendix \ref{subsection:easy_auc_m} to support our arguments. When we use the min-max formulation~(\ref{opt:spp}) to explain these issues, we will make some simplification. In particular, we will use the optimal value of $a, b,\alpha$ given $\w$, i.e., $a = a(\w) := \E[h_\w(\x)|y=1], b = b(\w) := \E[h_\w(\x)|y=-1], {\alpha =  1+ b - a}$, where $a, b$ can be interpreted as the mean prediction score on positive data and negative data, respectively {(please refer to Appendix \ref{section:optimal_a_b_alpha} for a derivation)}. The same trick will be used to illustrate the benefit of the AUC Margin loss. 

\noindent{\bf Adverse Effect on Easy Data}.  To illustrate this, let us consider a scenario: the current model parameter is given by $\w$ and there comes a positive and negative data pair $(\x, y=1), (\x', y'=-1)$. Suppose these data are easy examples meaning that the prediction $h_\w(\x)$ is large and $h_\w(\x')$ is small such that $h_\w(\x) - h_\w(\x')>1$. By taking the stochastic gradient descent update of the square loss $\ell(h_\bw (\x) - h_\bw(\x'))= (1 -h_\bw (\x) + h_\bw(\x') )^2$, we have the updated model given by  $\w_+ = \w- \eta 2(1-h_\bw (\x) + h_\bw(\x') )( -\x +\x')$, where $\eta>0$ is  a step size. Since $1-h_\bw (\x) + h_\bw(\x') <0$, the model parameter $\w$ will move towards the negative direction of the positive data $\x$ and the positive direction of the negative data $\x'$. As a result, the new model $\w_+$ tends to push the score $h_{\w_+}(\x)$ on the positive data smaller and the score  $h_{\w_+}(\x')$ on the negative data larger, which makes its classification capability worse. A similar effect happens when we use the min-max objective~(\ref{opt:spp}) to conduct the update. We include the analysis in Appendix~\ref{app:easyminmax}. 

\noindent{\bf Sensitivity to Noisy Data}.  Next, we elaborate the issue of sensitivity to noisily labeled examples. To this end, we consider a scenario: the current model parameter is given by $\w$ and there comes a positive and negative data pair $(\x, y=1, \hat y=-1), (\x', y'=-1, \hat y'=1)$, where $y, y'$ denote the true labels of $\x, \x'$ that are not revealed, respectively, and $\hat y=-1, \hat y'=1$ denote the noisy labels. 
Again, assume the prediction $h_\w(\x)$ is large and $h_\w(\x')$ is small. The SGD update of the model parameter $\w$ based on the min-max objective is given by 
\[
\w_+ = \w - 2\eta \{(1-p)(h_\bw (\x') - a- \alpha)\x' + p(h_\bw (\x)- b+\alpha)\x\}.
\]  By plugging the optimal values of $a, b, \alpha$ given $\w$, i.e.,  $\alpha = 1+ b - a$ and $a = \EX[h_\w(\x)|y=1],  b = \EX[h_\w(\x')|y'=-1]$, we can see that the term in the update of $\w$ that involves $\x$ is $-2\eta p(h_\bw (\x) + 1 - \EX[h_\w(\x)|y=1])\x$, and that involves $\x'$ is $-2\eta p(h_\bw (\x') - 1 - \EX[h_\w(\x')|y'=1])\x'$. Then it is clear to see that when $h_\w(\x)$ is large enough such that $h_\bw (\x) + 1 - \EX[h_\w(\x)|y=1]>0$, the update of $\w$ will move to the negative direction of the truly positive data $\x$, and similarly it will move to the positive direction of the truly negative data $\x'$ when $h_\w(\x')$ is small enough. 

\subsection{The Proposed AUC Margin Loss}
\vspace{-0.05in}
To alleviate the two issues of  the AUC square loss,  we propose a new margin-based surrogate loss. The new surrogate loss is a direct modification of the square loss to alleviate the two issues. 
To motivate the new AUC margin loss, we reformulate the AUC square loss as following (please refer to Appendix \ref{section:reformulation_auc_square_loss} for a derivation):  
\begin{align}\label{eqn:sq}
\hspace*{-0.05in}&A_{\text{S}}(\w)= \EX[(1 - h_\w(\x) +  h_\w(\x'))^2|y=1, y'=-1]\notag\\
  & = \underbrace{\EX[(h_\w(\x) - a(\w))^2|y=1]}\limits_{A_1(\w)}\\
  &+  \underbrace{\EX[(h_\w(\x') - b(\w))^2|y'=1]}\limits_{A_2(\w)}+  \underbrace{(1- a(\w) +  b(\w))^2}\limits_{A_3(\w)}\notag\\
    & =A_1(\w) + A_2(\w) + \max_{\alpha}\{2\alpha(1- a(\w) +  b(\w)) - \alpha^2\}\notag,
\end{align}
where $a(\w) = \EX[h_\w(\x)|y=1], b(\w) = \EX[h_\w(\x')|y'=1]$, and in the second equality we use the fact $s^2=\max_{\alpha}2\alpha s - \alpha^2$.  The three terms $A_1(\w), A_2(\w), A_3(\w)$ have meaningful interpretations.  In particular, minimizing $A_1(\w), A_2(\w)$ aim to minimize the variance of prediction scores on positive data and negative data, respectively;  minimizing the $A_3(\w)$ aims to push the mean prediction scores of positive and negative examples to be far away. However, the square function in the last term makes it suffer from the two aforementioned issues. Our solution is to use a squared hinge function to replace $A_3(\w)$, which is widely used in margin-based SVM classifiers. In particular, we replace $A_3(\w)$ by $\max_{\alpha\geq 0}\{2\alpha(m- a(\w) +  b(\w)) - \alpha^2\} = (m-a(\w) + b(\w))_+^2$, where $m$ is a hyper-parameter that specifies desired margin between $a(\w)$ and $b(\w)$. 
Hence, {\bf our new AUC margin loss is defined by} 
\begin{align}\label{eqn:AUCM}
&A_{\text{M}}(\w)= A_1(\w) + A_2(\w)\\
  &+ \max_{\alpha\geq 0}2\alpha (m- a(\w) +  b(\w)) - \alpha^2\nonumber.
\end{align}
Without the non-negative constraint on $\alpha$, the loss becomes the square loss  with a tunable margin parameter $m$. 

\noindent{\bf Benefits of the AUC Margin Loss}. 
We first show that the above objective is equivalent to a min-max objective. 
\begin{theorem}\label{thm:AUCM}
Minimizing the AUC margin loss~(\ref{eqn:AUCM}) is equivalent to  the following min-max optimization: 
\begin{align}\label{opt:sppM}
\min_{\w\in\R^d \atop (a,b)\in\R^2}\max_{\alpha\geq 0}\EX_{\z}\left[F_{\text{M}}\left(\w,a,b,\alpha;\z\right)\right], \quad \text{where}
\end{align}
 \begin{align}\label{eqn:AUCFM}&F_{\text{M}}(\w,a,b,\alpha;\z)=(1-p)\left(h_\w(\x)-a\right)^2\mathbb{I}_{[y=1]}\\
&+p(h_\w(\x)-b)^2\mathbb{I}_{[y=-1]}-p(1-p)\alpha^2\notag\\
&+2\alpha\left(p(1-p)m+ p h_\w(\x)\mathbb{I}_{[y=-1]}-(1-p)h_\w(\x)\mathbb{I}_{[y=1]}\right)\notag. 
\end{align}
\end{theorem}

\noindent
We highlight that $\min_{a,b}\max_{\alpha\geq 0}\EX_{\z}\left[F_{\text{M}}\left(\w,a,b,\alpha;\z\right)\right]=p(1-p)A_\textM(\w)$. 
Please see proof in Appendix \ref{section:proof_of_auc_margin_equivalence}.

\noindent{\bf Robust to Easy Data.} Based on the above min-max formulation, let us first elaborate the benefits of the new loss that alleviate the two issues of the AUC square loss. First, let us consider how the non-negative constraint $\alpha\geq 0$ helps alleviate the adverse effect when trained with easy data. Following the same logic as before, we compute the gradient of $F_{\text{M}}(\w, a, b, \alpha)$ by 
\begin{align*}
\nabla_\w F_{\text{M}}(\w, a, b, \alpha; \z) =&
2(1-p)\x \I_{[y=1]} \cdot (h_\w(\x) - a - \alpha)
\\
&
+ 2p\x \I_{[y=-1]} \cdot (h_\w(\x) - b + \alpha) .
\end{align*}
Different from the square loss, the optimal $\alpha$ given $\w$ is $\alpha = m + b(\w) - a(\w)$ if $m+b(\w) - a(\w)\geq 0 $, and $\alpha = 0$ if $m+b(\w) - a(\w)< 0 $, where $a(\w) = \E[h_\w(\x)|y=1], b(\w) = \E[h_\w(\x)|y=-1]$. When the model is good enough, i.e., $m + b(\w) - a(\w)<0$ meaning that the mean prediction scores of positive data is larger than the mean prediction scores of negative data by a margin $m>0$, then the gradient becomes $\nabla_\w F_{\text{M}}(\w, a, b, \alpha; \z) = 2(1-p)\x \I_{[y=1]} \cdot (h_\w(\x) - a) +  2p\x \I_{[y=-1]} \cdot (h_\w(\x) - b)$.  Taking a stochastic gradient decent update for $\w$ will only push the prediction score of the sampled data to be close to their mean score. When the model is poor, i.e., $m + b(\w) - a(\w)>0$,  the gradient becomes $\nabla_\w F_{\text{M}}(\w, a, b, \alpha; \z) = 2(1-p)\x \I_{[y=1]} \cdot (h_\w(\x) - m - b(\w)) +  2p\x \I_{[y=-1]} \cdot (h_\w(\x) + m - a(\w))$.  Since the model is poor in this case, it is likely that $h_\w(\x) - m - b(\w)<0$ for a positive data $\x$, and $h_\w(\x) + m - a(\w)>0$ for a negative data $\x$. As a result, taking a stochastic gradient decent update for $\w_+ = \w- \eta \nabla_\w F_{\text{M}}(\w, a, b, \alpha; \z)$ will likely move the model in the right direction pushing the prediction score of positive data larger, and that of negative data smaller. 

\noindent{\bf Robust to Noisy Data}. Next, let us elaborate  how adding a tunable margin parameter $m$ can help alleviate the sensitivity to noisy data. Similar to the AUC square loss, the update in the noisy data case is given by 
\[
\w_+ = \w - 2\eta \{(1-p)(h_\bw (\x') - a- \alpha)\x' + p(h_\bw (\x)- b+\alpha)\x\},
\]
where $\x'$ is a true negative data but labeled as positive and $\x$ is a true positive data but labeled as negative. Let us consider the case that model is not good enough such that the optimal value of  $\alpha = m + b(\w) - a(\w)$. Then the term in the update of $\w$ that involves the true positive data $\x$ is $-2\eta p(h_\bw (\x) + m - \EX[h_\w(\x)|y=1])\x$, and that involves the true negative data $\x'$ is $2\eta p(m +  \EX[h_\w(\x')|y'=1] - h_\bw (\x'))\x'$. Note that even when $h_\w(\x)$ is large and $h_\w(\x')$ is small such that the model $\w_+$ is moving in the wrong direction, by tuning $m$ to a smaller value, we can  ensure that the movement into the wrong direction is much reduced. Hence, adding the tunable margin parameter $m$ can alleviate the sensitivity to the noisy data.

\subsection{DAM with the AUC Margin Loss}
As seen from Theorem~\ref{thm:AUCM}, the AUC margin loss is equivalent to a min-max optimization problem, that is similar to that of the AUC square loss. Hence, any stochastic algorithms proposed for solving the min-max objective of the AUC square loss can be easily adapted to solving the min-max objective of the AUC margin loss. In particular, for any update on the dual variable $\alpha$, we follow by a projection step that projects $\alpha$ into non-negative orthant. In this paper, we employ the proximal epoch stochastic method (named PESG) proposed in \cite{guo2020fast} to update variables $\w, a, b, \alpha$. 
To present the algorithm,  we use a notation $\mathbf v=(\w, a, b)$ to denote all primal variables.  The key steps are presented in Algorithm~\ref{alg:primal_dual_auc}. In the algorithm, $\lambda$ denotes the standard regularization parameter (i.e, weight decay parameter), $\gamma>0$ is an algorithmic regularization parameter that can help improve the generalization, $\mathbf v_{\text{ref}}$ is  a reference solution that is updated periodically by using the accumulated average  of $\mathbf v_t$ in the previous stage (before decaying learning rate). We refer the readers to~\cite{liu2019stochastic, guo2020fast} for more discussion and convergence analysis of this algorithm. 

\setlength{\textfloatsep}{5pt}

\begin{algorithm}[t]
\caption{PESG for optimizing the AUC margin loss}{{\bf Require}: $\eta, \gamma, \lambda, T$}
\label{alg:primal_dual_auc}
\begin{algorithmic}[1]
\STATE Initialize $\mathbf v_1, \alpha_1\geq 0$

\FOR{$t=1, \ldots, T$}

\STATE Compute $\nabla_{\mathbf v} F_{\text{M}}(\mathbf v_t, \alpha_t; \z_t)$ and $\nabla_\alpha F_{\text{M}}(\mathbf v_t, \alpha_t; \z_t)$.

\STATE Update primal variables
\[
\mathbf v_{t+1} = \mathbf v_{t} - \eta (\nabla_{\mathbf v} F_{\text{M}}(\mathbf v_t, \alpha_t; \z_t)+ \gamma (\mathbf v_t-\mathbf v_{\text{ref}})) - \lambda \eta\mathbf v_t 
\]
\STATE Update $\alpha_{t+1}=  [\alpha_{t} + \eta \nabla_\alpha F_{\text{M}}(\mathbf v_t, \alpha_t; \z_t)]_+$. 
\STATE Decrease $\eta$ by a  factor and update $\mathbf v_{\text{ref}}$ periodically  

\ENDFOR
\end{algorithmic}
\end{algorithm}
%
\noindent{\bf A Two-stage Framework for DAM.}
From our preliminary studies on deep AUC maximization, we observe that directly optimizing the AUC margin loss can easily handle the recognition tasks on simple datasets, e.g., CIFAR. However, it shows some difficulties on complex tasks, e.g., CheXpert, Melanoma. We conjecture that the feature extraction layers learned by directly optimizing AUC from scratch are not as good as optimizing the standard cross-entropy loss on these difficult data.  Inspired by recent works on two-stage methods, e.g., \cite{Kang2020Decoupling}, we also employ a two-stage framework \textbf{on difficult medical image classification tasks} that includes a \textit{pre-training} step that minimizes the standard cross-entropy loss, and an \textit{AUC maximization} step that maximizes an AUC surrogate loss of the pre-trained CNN for learning all layers with the last classifier layer randomly initialized.

\section{Empirical Studies}
\label{exp_section}
In this section, we present extensive empirical studies on the proposed robust DAM method with the AUC margin loss. First, we present results on some benchmark datasets and then we present the results on four medical image classification tasks. The code for reproducing the results of our method in this paper can be found here~\cite{DeepAUC}.

\subsection{Performance on Benchmark datasets}
\label{section:bechmark_results}
 For benchmark datasets, we construct imbalanced Cat\&Dog (C2), CIFAR-10 (C10), CIFAR-100 (C100), STL-10 (S10)~\cite{elson2007asirra, krizhevsky2009learning, coates2011analysis} following instructions by \cite{liu2019stochastic}. Specifically, we first randomly split the training data by class ID into two even portions as the positive and negative classes, and then we randomly remove some samples from the positive class to make it imbalanced. We keep the testing set untouched. We refer to imbalance ratio (\textbf{imratio}) as the ratio of \# of positive examples to \# of all examples. Statistics of these datasets are presented  in Appendix \ref{section:dataset_description}. 

We experiment with two network structures,  i.e., DenseNet121 (\textbf{D}) (\cite{huang2017densely}) and ResNet20 (\textbf{R}) (\cite{he2016deep}) with ELU activation functions. We explore the imbalance ratio = 1\%, 10\%, and  use a 9:1 train/val split to conduct cross-valuation for tuning parameters. We compare DAM using our AUC margin loss (AUC-M) with three baselines, DAM using AUC square loss (AUC-S),  and DL with two other popular loss functions i.e.,  cross-entropy loss (CE) and focal loss (Focal) trained by SGD.  We use the ${\hat \alpha}$-balanced Focal loss $-{\hat \alpha}(1-p_t)^{\hat \gamma} \log(p_t)$, and tune its parameter ${\hat \alpha}, \hat \gamma$ from [0.25, 0.5, 0.75] and [1,2,5] on the validation set, respectively. For DAM, we tune $\gamma$ in [1/100, 1/300, 1/500, 1/700, 1/1000]. For AUC-M loss, we tune margin parameter $m$ in [0.1, 0.3, 0.5, 0.7, 1.0]. For optimization, we run 100 epochs with a stagewise learning rate: initial value of 0.1 and decaying at $50\%$ and $75\%$ of the total number of training epochs for all experiments. We use a weight decay, i.e., $\lambda$, as $1e$-$4$ for all methods. The batch size is set to 128 on all datasets except for S10, which is set to 32 due to smaller data size. For each method, we run the experiment with five different random training sets (by randomly removing some positive examples with different random seeds), and evaluate on the same testing set by comparing the averaged testing AUC scores. {We also found that using a L2 normalization of the predication scores in a mini-batch is helpful. We refer to this normalization as \textbf{Batch Score Normalization} (BSN). Hence, in the following experiments we use the BSN before computing both the AUC-S and AUC-M losses. Please refer to section \ref{sec:ablation_study_bsn} for an ablation study on comparing with and without BSN.}

The results for DenseNet121/ResNet20 with imratio=1\% are reported in Table \ref{tab:results_benchmark}. We include the results for imratio=10\% to the Appendix \ref{section:more_benchmar_results}. Overall, we observe that the AUC-M and AUC-S perform much better than non-AUC-based losses in most cases. Comparing AUC-M with AUC-S, we can see that AUC-M performs better in most cases, especially in the extremely imbalanced setting with imratio=$1\%$.

We also conduct some ablation studies on the benchmark datasets to demonstrate the robustness of the proposed AUC-M loss in comparison with AUC-S loss for DAM with added easy and noisy data, and the effectiveness of non-negative constraint on $\alpha$. The results are included in Section~\ref{sec:ab}.  
\begin{table}[t]
\centering
\caption{{ Testing AUC on benchmark datasets with imratio=1\%.} }
\label{tab:results_benchmark}
\scalebox{0.95}{
\begin{tabular}{ccccc}
\hline
\textbf{Dataset} & \textbf{CE} & \textbf{Focal} & \textbf{AUC-S} & \textbf{AUC-M}    \\ \hline
C2  (D)            & 0.718$\pm$0.018          & 0.713$\pm$0.009    & 0.803$\pm$0.018 & \textbf{0.809$\pm$0.016}       \\ 
C10 (D)             & 0.698$\pm$0.017          & 0.700$\pm$0.007    & 0.745$\pm$0.010          & \textbf{0.760$\pm$0.006} \\ 
S10 (D)           & 0.641$\pm$0.032          & 0.660$\pm$0.027    & 0.669$\pm$0.070          & \textbf{0.703$\pm$0.030}  \\ 
C100 (D)          & 0.588$\pm$0.011          & 0.591$\pm$0.017    & 0.607$\pm$0.010          & \textbf{0.614$\pm$0.016} \\ \hline
C2  (R)     & 0.730$\pm$0.028 & 0.724$\pm$0.020 & 0.748$\pm$0.007 & {\bf0.756$\pm$0.017}    \\
C10 (R)       & 0.690$\pm$0.011 & 0.681$\pm$0.011 & 0.702$\pm$0.015 & {\bf0.715$\pm$0.008}    \\
S10 (R)      & 0.641$\pm$0.021 & 0.634$\pm$0.024 & 0.645$\pm$0.029 & {\bf0.659$\pm$0.020}   \\
C100 (R)      & 0.563$\pm$0.015 & 0.565$\pm$0.022 & 0.587$\pm$0.017 & {\bf0.596$\pm$0.016}  \\ \hline
\end{tabular}}
\end{table}

\subsection{Medical Image Classification Tasks}
Below, we present results on  four difficult medical image classification tasks, namely classification  of  X-ray  images  for detecting  chest  diseases,  classification of  images  of  skin  lesions  for detecting melanoma,  classification  of mammograms  for  breast  cancer screening,   and  classification  of  microscopic  images  for identifying tumor  tissue.  A summary of these tasks and their data is reported in Table~\ref{tab:meddata}.

\subsubsection{CheXpert Competition}
\label{section:chexpert}
{\bf CheXpert} competition is a medical AI competition organized by Stanford ML group~\cite{irvin2019chexpert}, which released a large-scale Chest X-Ray dataset for detecting chest and lung diseases~\cite{irvin2019chexpert}. The training data consists of 224,316 high-quality X-ray images from 65,240 patients. The validation dataset consists of 234 images from 200 patients.  The testing data has images for 500 patients, which is not released to the public and is maintained by the organizer for final evaluation. The training images were annotated by a labeler to automatically detect the presence of 14 observations in radiology reports, capturing uncertainties inherent in radiography interpretation.  The validation images were manually annotated by 3 board-certified radiologists. The testing images were annotated by a consensus of 5 board-certified radiologists. The average resolution of CheXpert images is 2828x2320 pixels, which is about 6 times larger than ImageNet. The competition requires participants to submit the trained models for evaluation of the AUC score on predicting 5 selected diseases, i.e., Cardiomegaly, Edema, Consolidation, Atelectasis, Pleural Effusion. These tasks have an average imratio of 20.21\%.  They also reported another metric that compares the model's performance with 3 radiologists' predictions for reference.

\begin{table}[t]
\centering
\caption{Summary of Medical Classification Tasks.}
\scalebox{0.95}{
\begin{tabular}{cccc}
\hline
\textbf{Dataset}  & Image Domain& Imratio & \# Training  \\ \hline
CheXpert &Chest X-ray & 20.21\% &224,316 \\
Melanoma &Skin Lesion  & 7.1\%& 46,131\\
DDSM+   &Mammogram &13\%& 55,000\\ 
PatchCamelyon  & Microscopic & 1\%& 148,960 \\ \hline
\end{tabular}}
\label{tab:meddata}
\end{table}

\textbf{Model Pre-training}. To tackle the uncertain data in CheXpert, we adopt a label smoothing method similar to that in works~\cite{pham2020interpreting}. 
We choose five networks: DenseNet121, DenseNet161, DensNet169, DensNet201 and Inception-renset-v2\cite{huang2017densely,szegedy2016inception}. With limited resources, we scale the resolution of all raw images to 320x320. For data augmentation, we use random rotation, random translation and random scaling. For \textit{pre-training} step, we optimize CE loss by Adam on the 5 classification tasks with weight decay parameter of 1e-5. The total training time is 2 epochs with a batch size of 32 and initial learning rate of 1e-5. 
In the second step of AUC maximization, we replace the last classifier layer trained in the first step by random weights and use our DAM method to optimize the last classifier layer and all previous layers. 
We tune  $\gamma$ in \{1/300, 1/500, 1/800\}, set weight decay $\lambda$ to 0,  set the initial learning rate to 0.1 and decrease the learning rate at 2000, 8000 iterations by 3 times, run a total of 2 epochs for Algorithm \ref{alg:primal_dual_auc}.

\textbf{Competition Results}. Our final submission is the ensemble of five models trained by DAM with the AUC-M loss for each disease. On Aug 31, 2020, we submitted our models to CheXpert and we achieved a mean testing AUC score of \textbf{0.9305}, which is currently ranked at \textbf{1st place} over all submissions. The leaderboard is shown in \cite{chexpert_2019}, where our submission is named as DeepAUC-v1 (ensemble). We also compare our results with other methods in Table \ref{table:chexpert_test}, where Hierarchical Learning \cite{pham2020interpreting} utilizes domain knowledge to pre-define a disease hierarchy used for conditional training,  YWW~\cite{ye2020weakly} utilizes weakly-supervised lesion localization technique through a novel Probabilistic-CAM (PCAM) pooling operator to improve the model training. All these solutions are trained by CE loss. Our AUC-based solution surpasses these solutions and it is also better than 2.8 out of 3 radiologists (NRBC) for 5 selected diseases on average as in Table~\ref{table:chexpert_test}. Finally, we noticed that a recent work that optimizes AUC square loss for DAM on CheXpert only achieves a mean testing AUC score of 0.922~\cite{guo2020fast}. 

\begin{table}[t]
\centering
\caption{Averaged Testing AUC Scores on CheXpert. NBRC means the \# of radiologists out of 3 are beaten by AI algorithms.}
\scalebox{0.95}{
\begin{tabular}{llll}
\hline
Model & AUC  & NRBC & Rank   \\ \hline
\textbf{Stanford Baseline}~\cite{irvin2019chexpert} & 0.9065 & 1.8 & 85 \\ 
\textbf{YWW}~\cite{ye2020weakly} &  0.9289 & 2.8 & 5 \\ 
\textbf{Hierarchical Learning}~\cite{pham2020interpreting}& 0.9299 & 2.6 & 2 \\ 
\textbf{DAM (Ours)} & \textbf{0.9305} &  \textbf{2.8} & \textbf{1} \\ \hline
\end{tabular}}
\label{table:chexpert_test}
\end{table}

\subsubsection{Melanoma Classification}
Melanoma is a skin cancer, which is the major cause for skin cancer death~\cite{miller2006melanoma}. We conduct empirical studies on the Kaggle Melanoma dataset \cite{rotemberg2020patient}, which is released through a  Kaggle competition. The data  is split into 33,126 training images with 584 malignant melanoma images (imbalance ratio=1.76\%) and 10,892 testing images with an unknown number of melanoma images. Further, the testing set is split into public testing set and private testing set at 30\%/70\% ratio by patient ID. The public testing set (noting that their ground-truth labels are not revealed) is used to rank participating teams at the early stage. The private testing set is used to evaluate the  participating teams for the final ranking. The public AUC score is  updated daily but private AUC score is released after the end of competition.

\textbf{Data preparations}. The raw dataset has various sizes of images, e.g., 6000x4000, 1920x1080. We resize all images to lower resolutions 
due to limited computational resources. To evaluate the model locally, we follow \cite{ChrisTFrecords} to construct a 5-fold Stratified Leak-Free version cross-validation by 8:2 train/valid split. The data split follows two rules: 1) images from same patients are either put  in train set or in  validation set. 2) train and validation set have same imbalance ratio 1.76\%. In addition, we also utilize two external data sources to complement the provided data in train set: 1) 12,859 images from previous competitions, e.g., ISIC2017 and ISIC2018, and 2) 580 malignant melanoma images parsed from the website of The International Skin Imaging Collaboration~\cite{ISICdatasets}. We merge all data sources and finally obtain a training set of 46,131 images with an imbalance ratio of 7.1\%.

\textbf{Comparison with Baselines}. We first compare with three baselines as above, i.e., optimizing CE, Focal and AUC-S losses.  
We choose the family of EfficientNet~\cite{tan2019efficientnet} as the main network.  Data augmentation is very crucial in this competition, and we use a set of augmentations, e.g., horizontal flipping, rotating, scaling, shearing, coarse dropout following a public notebook \cite{ChrisTFrecords}.  In addition, we use the cyclical learning rate with a base learning rate \cite{smith2017cyclical} of 3e-5 and a maximum learning rate of 2.4e-4 and with  8 epochs for a full cycle. We use a weight decay of 1e-5. For  focal loss \cite{lin2017focal}, we tune ${\hat \gamma}$=\{1,2,5\}, ${\hat \alpha}$=\{0.25,0.5,0.75\} and report the best result. For non-AUC losses, we train a total of 16 epochs with batch size of 256.  
For DAM, we start optimization from the pretrained backbone trained by optimizing the CE loss. For AUC losses, we set $\gamma$ to 1/500 which is tuned by cross validation. For AUC Margin loss, we also tune $m=\{0.3, 0.5, 0.7, 1.0\}$. For experiments, we train 35 epochs in total with same batch size and initial learning rate of 0.01 decreasing by 2 times every 10 epochs using Algorithm \ref{alg:primal_dual_auc}. In addition, we find patient-level information (metadata) useful, e.g., age, sex, and location of imaged site. To utilize metadata, after training EfficientNet, we merge it with a 2-layer neural network (256x128) with a 0.5:0.5 weighted ratio, which is trained independently.  
The network structure is illustrated in Figure \ref{fig:kaggle_training_frameowork} in Appendix \ref{section:more_results_on_kaggle}. 

\begin{figure}[t]
\centering
\includegraphics[width=0.35\textwidth]{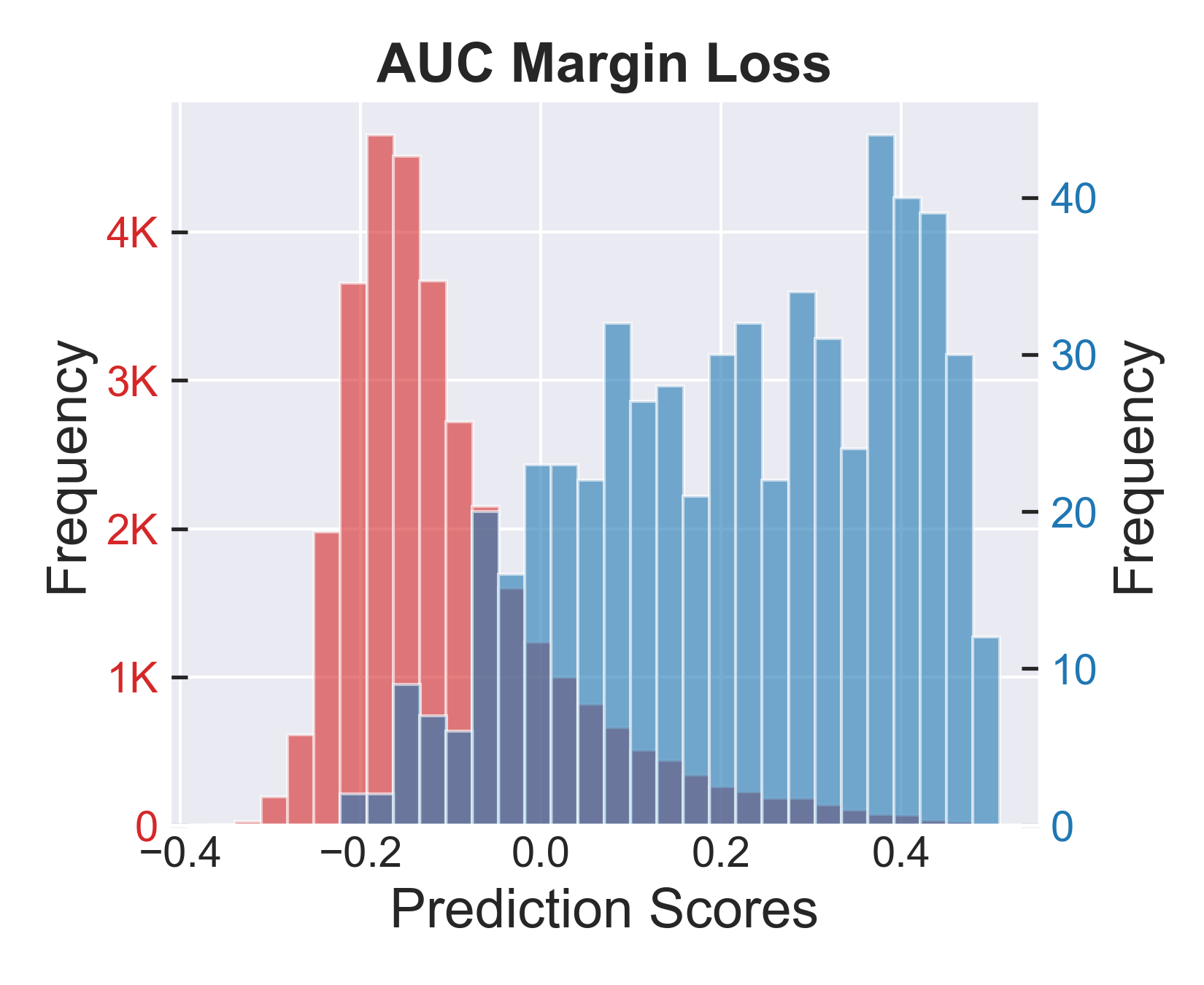}
\includegraphics[width=0.35\textwidth]{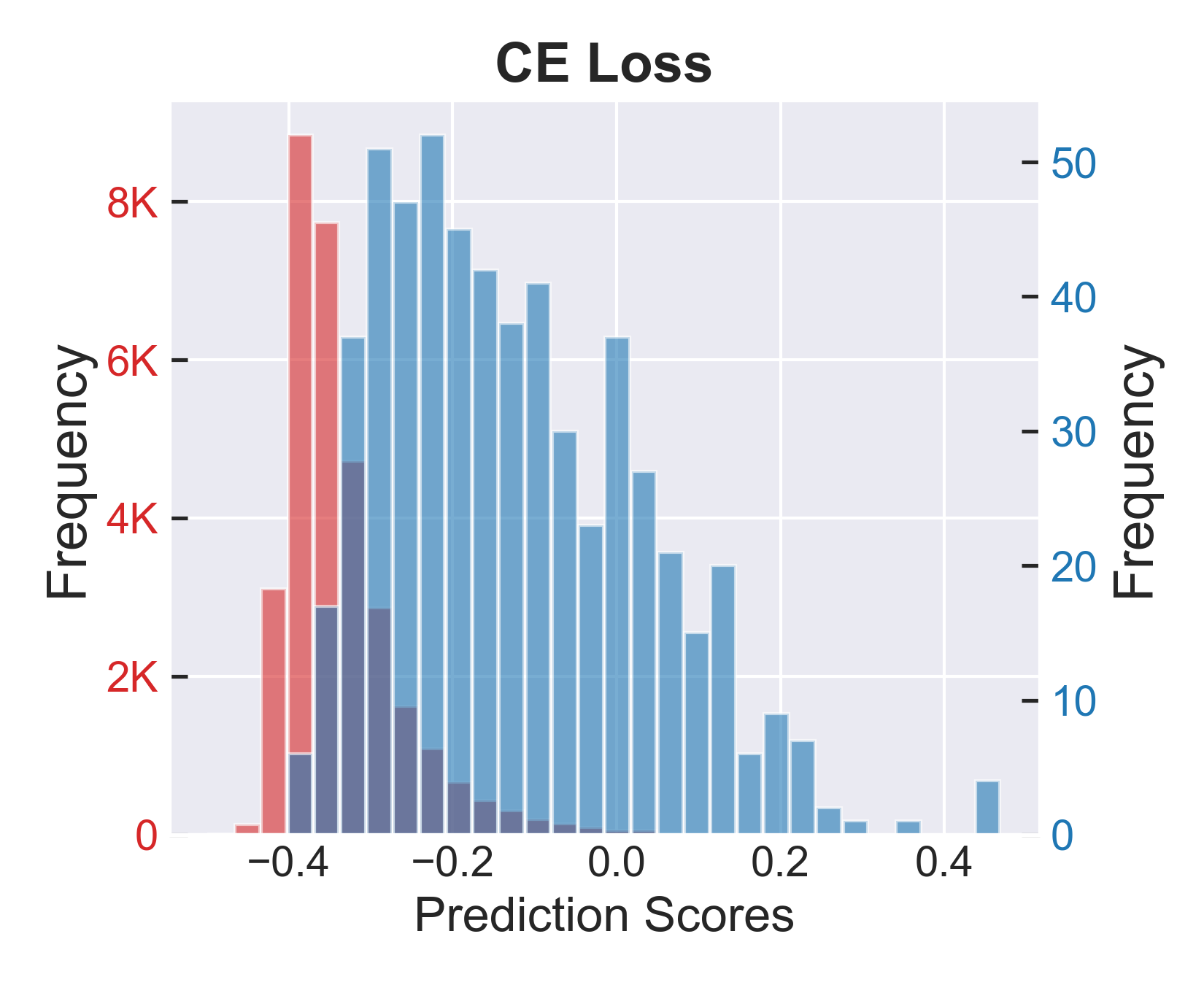}
\caption{Prediction histogram of positive (blue) and negative (red) samples for the models trained by AUC-M loss and CE loss on Melanoma training dataset.
}
\label{fig:melanoma_hist}
\end{figure}

The comparison between different methods for learning EfficientNet-B5 on resized images with a fixed resolution of $384\times 384$ is given in Table~\ref{tab:kaggle_results}. For each method, we report four numbers that represent performance on the public testing data (in early stage of competition) and private testing data (for final ranking) with/without test-time data augmentation (TTA)\cite{simonyan2014very}. 
We can see that DAM methods improve over the standard DL methods for minimizing CE and Focal losses. In addition, the AUC Margin loss is better than AUC Square loss. We also plot the histogram of predictions on training data of our best DAM method (AUC-M+Meta) compared with standard DL method with CE loss in Figure \ref{fig:melanoma_hist}. We can see that the predictions by the DAM method have two well-separated patterns corresponding to positive and negative data. In contrast,  the predictions by optimizing the CE loss is more mixed together.

\begin{table}[t!]
\caption{Comparison of Testing AUC on Melanoma dataset for Optimizing EfficientNetB5. TTA (30) means that predictions are averaged over 30 augmented copies of each image in test set.}
\centering
\scalebox{0.95}{
\begin{tabular}{cccccc}
\hline
\multicolumn{1}{l}{}        & \multicolumn{2}{c}{\textbf{wo/ TTA}} & \multicolumn{2}{c}{\textbf{w/ TTA(30)}} \\ \hline
\textbf{Loss}             & \textbf{Public}        & \textbf{Private}       & \textbf{Public}       & \textbf{Private}       \\ \hline
\multicolumn{1}{c|}{CE}  & 0.9391          & 0.9285            & 0.9447           & 0.9345             \\
\multicolumn{1}{c|}{Focal}  & 0.9412          & 0.9266            & 0.9424           & 0.9303             \\
\multicolumn{1}{c|}{AUC-S} & 0.9482          & 0.9332             & 0.9502           & 0.9364             \\
\multicolumn{1}{c|}{AUC-M} & \textbf{0.9497}     &  \textbf{0.9357}    &  \textbf{0.9503 }          &  \textbf{0.9393}            \\ \hline 
\multicolumn{1}{c|}{AUC-S (Meta)} & 0.9495          & 0.9358             & 0.9501           & 0.9409            \\
\multicolumn{1}{c|}{AUC-M (Meta)} & \textbf{0.9522}     &  \textbf{0.9380}    &  \textbf{0.9520 }          &  \textbf{0.9423}            \\ \hline 
\multicolumn{1}{c|}{Our Submission} & -    & -  &  \textbf{0.9685 }          &  \textbf{0.9438}  
\\ \hline
\end{tabular}}
\label{tab:kaggle_results}
\end{table}

\textbf{Competition Results.} 
For final submission towards this competition, we use an ensemble method. We train different nets including  EfficientNet (B3, B5, B6) and different resolutions , i.e.,  $256\times256, 384\times384, 512\times512, 768\times768$. Our final result is averaged over 10 models, which is also reported in Table~\ref{tab:kaggle_results}. Our method achieves AUC scores of \textbf{0.9685/0.9438} on public/private sets, which rank at \textbf{42nd/33rd} out of 3314 teams. To our best knowledge, this is also the first solution to optimize AUC in the competition. The winning team has an AUC score of 0.9490 on the private testing set~\cite{ha2020identifying}. We would like to emphasize that the winning team has used several useful tricks to improve the final result. In particular, they used an ensemble of 18 models and also used images at higher resolution of $896*896$. We expect these tricks can be also used for improving our results.  In terms of learning a single model, our DAM method has a higher AUC score of 0.9423 than their single model's AUC score of 0.9167 (e.g., model 7 under similar configurations, e.g.,  EfficientNetB5, 384x384, metadata~\cite{ha2020identifying}). { After the competition, we find the ensemble of EffecientNetB5($384*384$, AUC-M loss, metadata) and EffecientNetB6($512*512$, CE loss) achieves highest private AUC of \textbf{0.9505}}.

\subsection{Other Two Medical Classification Tasks}
Finally, we present results on two more medical classification tasks, i.e.,  classification of mammogram for breast cancer screening on DDSM+ data,  and classification of microscopic images for identifying tumor tissue on PathCamelyon Data. The DDSM+ data is a combination of two datasets namely DDSM and CBIS-DDSM \cite{bowyer1996digital,heath1998current}, which  consists of 55,000 mammographic images (224$\times$224) taken at lower doses than usual X-rays for training with imratio of 13\% and 13,900 images for testing with imratio of 4\%. The PathCamelyon dataset consists of 294,912 color images (96$\times$96) extracted from histopathologic scans of lymph node section for training and 32,768 images for testing with balanced class ratio~\cite{veeling2018rotation, bejnordi2017diagnostic}. For second task, we manually construct an imbalanced dataset with imratio of 1\% following section \ref{section:bechmark_results}. For experiments, we train DenseNet121 and use batch size of 32  for DDSM+ and 64 for PatchCamelyon. For non-AUC losses, we train models using Adam with weight decay of 1e-5 for 5 epochs. We tune learning rate \{1e-1 $\sim$ 1e-5\} on validation set sampled from 10\% training data. For focal loss, we tune ${\hat \gamma}$=\{1,2,5\}, ${\hat \alpha}$=\{0.25,0.5,0.75\}. For AUC losses, we start from pretrained model of last iteration by CE loss and train a total of 1 epoch. We tune learning rate \{1e-1, 1e-2, 1e-3\}, $\gamma$=\{1/300, 1/500, 1/800\} and set $\lambda=0$. For AUC-M, we tune m=\{0.3, 0.5, 0.7, 1.0\}. We report the best results for each method in table \ref{tab:more_exp_different_domains1}. The results indicate that AUC-M performs consistently better than baseline methods on these two datasets.

\begin{table}[t]
\centering
\caption{Testing AUC of two medical datasets on DenseNet121.}
\scalebox{0.95}{
\begin{tabular}{ccccc}
\hline
\textbf{Data (imratio)} & \textbf{CE} & \textbf{Focal} & \textbf{AUC-S} & \textbf{AUC-M} \\ \hline
DDSM+ (13\%) & 0.9392 & 0.9495 & 0.9469 & \textbf{0.9544} \\ 
PatchCamelyon (1\%) & 0.8394 & 0.8556 & 0.8703 & \textbf{0.8896} \\ \hline
\end{tabular}}
\label{tab:more_exp_different_domains1}
\end{table}

\section{Ablation Studies}
\subsection{Batch Score Normalization (BSN)}
\label{sec:ablation_study_bsn}
\begin{figure}[h!]
\centering
\includegraphics[width=0.35\textwidth]{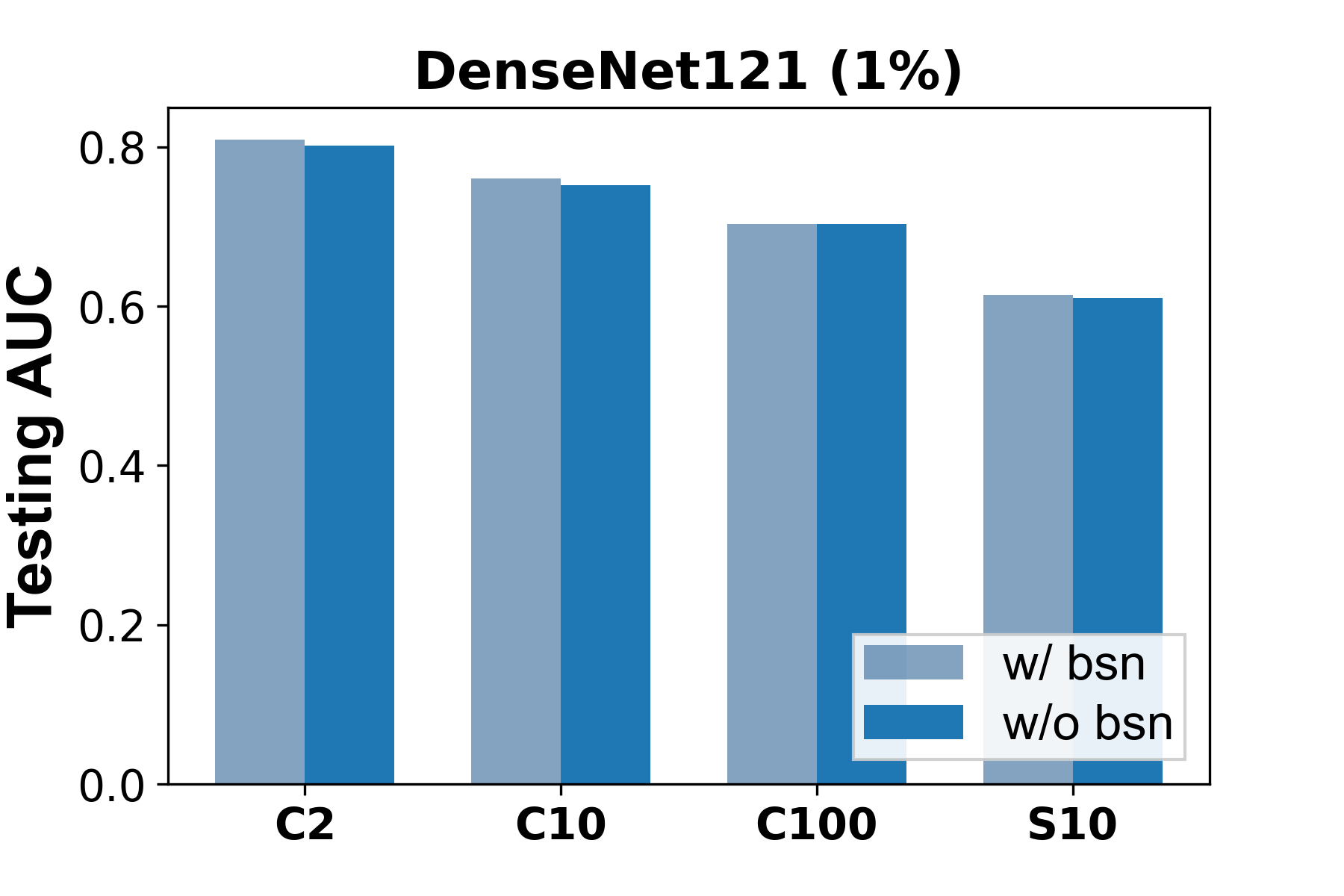}
\includegraphics[width=0.35\textwidth]{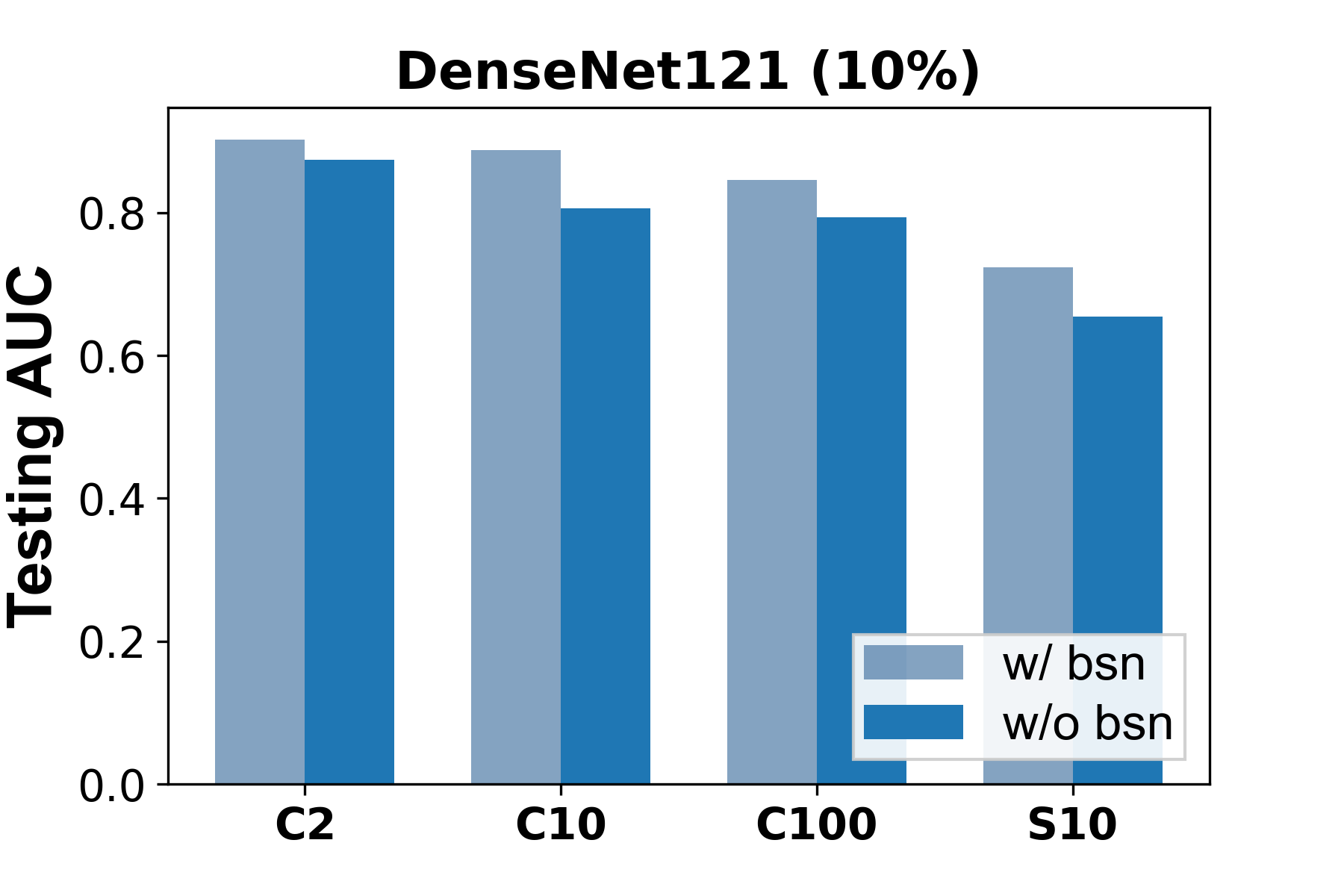}
\caption{Ablation Study on Batch Score Normalization. }
\label{fig:bsn}
\end{figure}

We run experiments with DenseNet121 on four benchmark datasets with two imbalance ratio, e.g., 1\%, 10\% with and without applying batch score normalization. The results are shown in Figure~\ref{fig:bsn}. We can see that applying the BSN can improve the performance. 

\subsection{AUC-Margin Loss}\label{sec:ab}
\begin{figure}[h]
\centering
\includegraphics[width=0.24\textwidth]{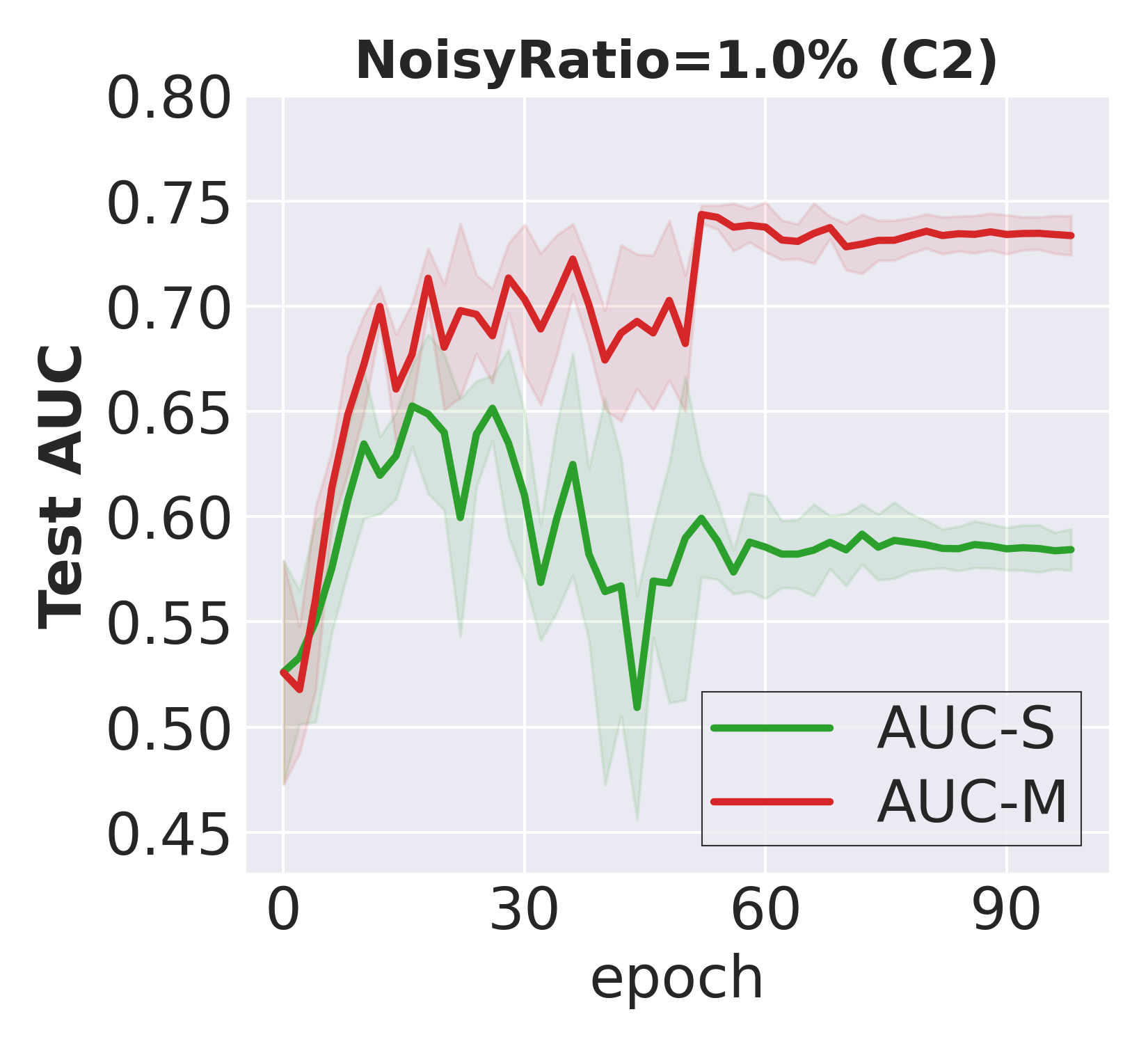}
\includegraphics[width=0.24\textwidth]{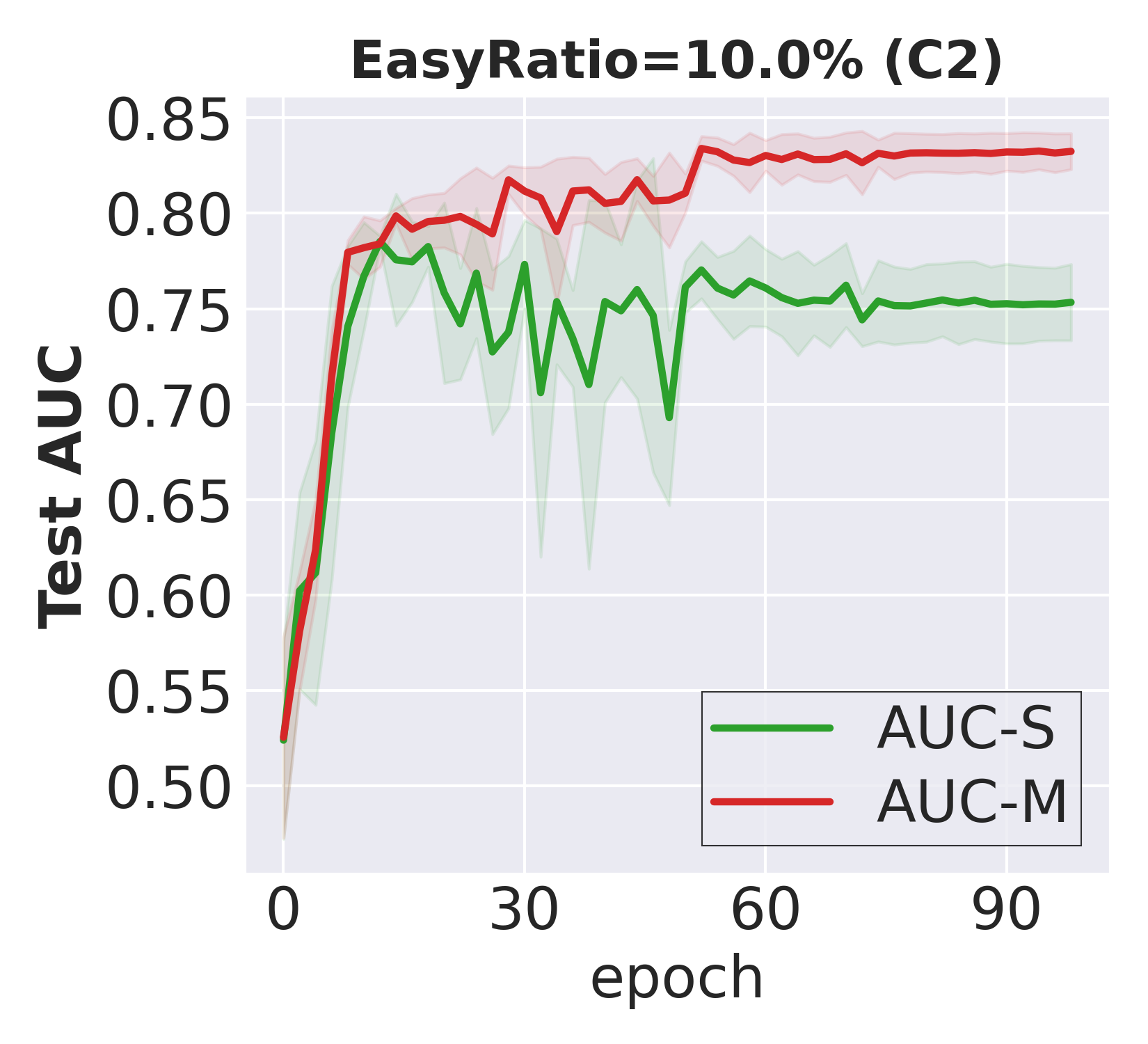}
\includegraphics[width=0.24\textwidth]{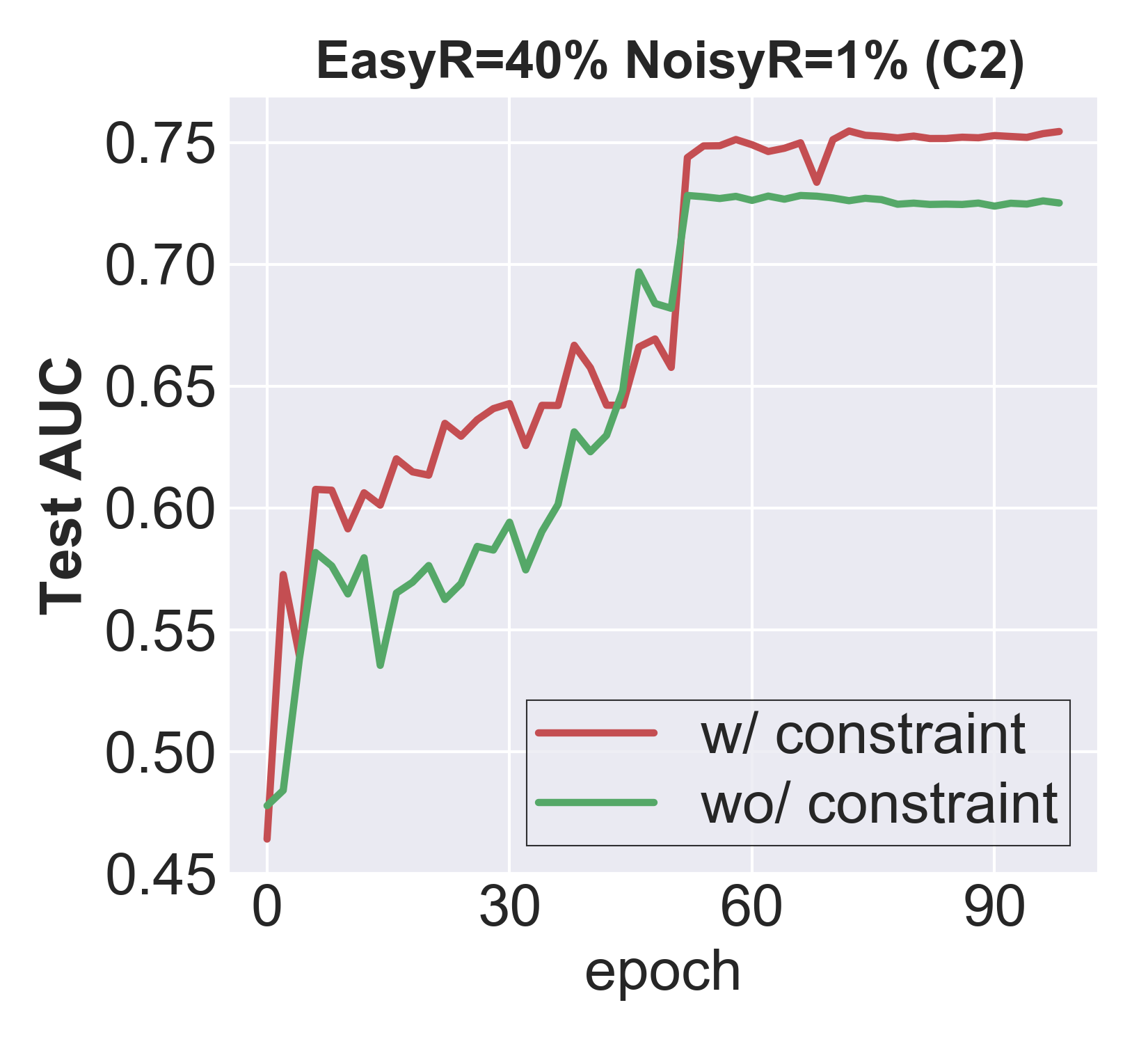}
\includegraphics[width=0.24\textwidth]{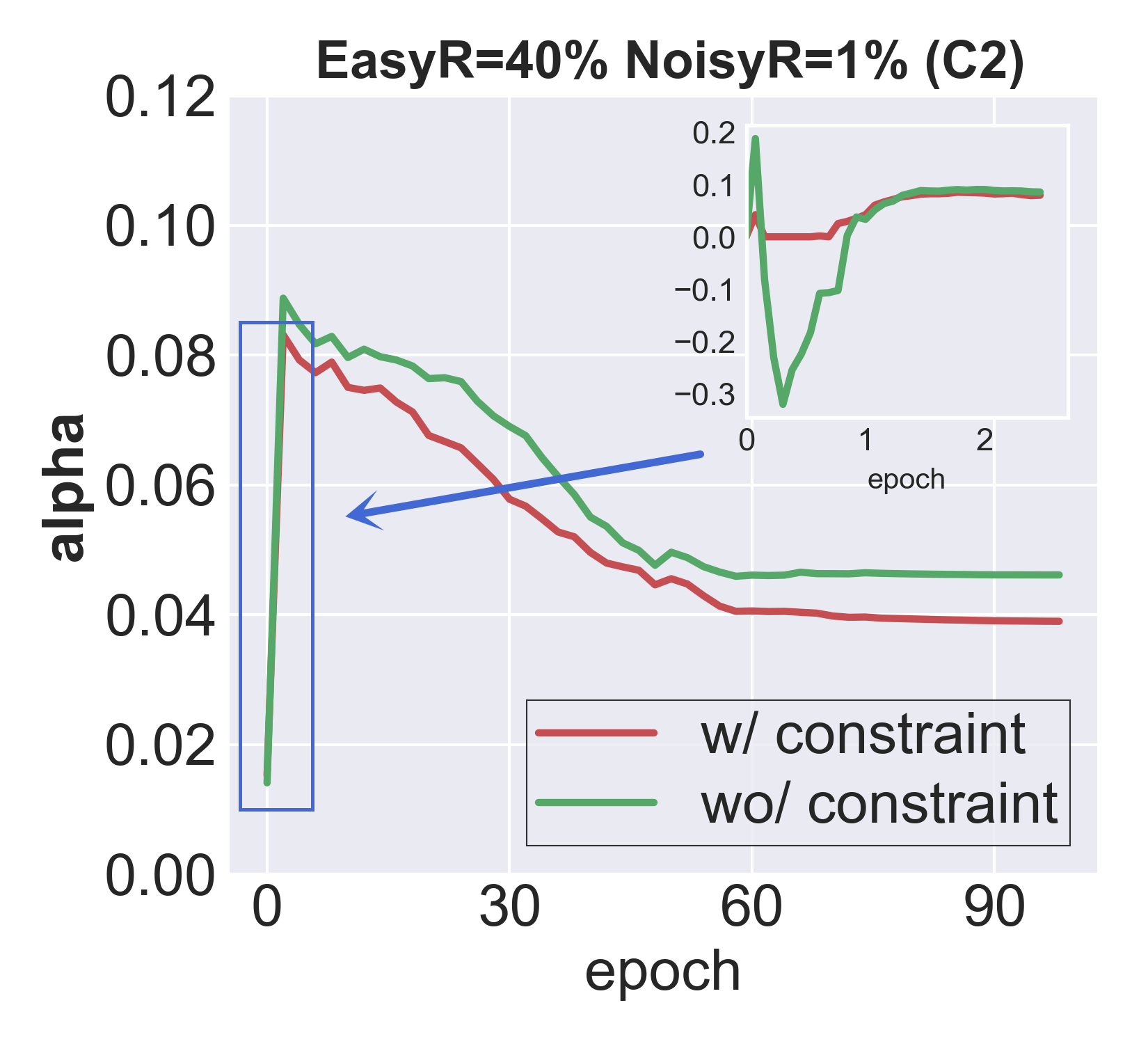}
\caption{First two plots: comparison when adding noisy and easy samples. Last two plots: comparison between with/without $\alpha\geq 0$.}
\label{fig:noisy_easy_example}
\end{figure}

\begin{figure}[h]
\centering
\includegraphics[width=0.24\textwidth]{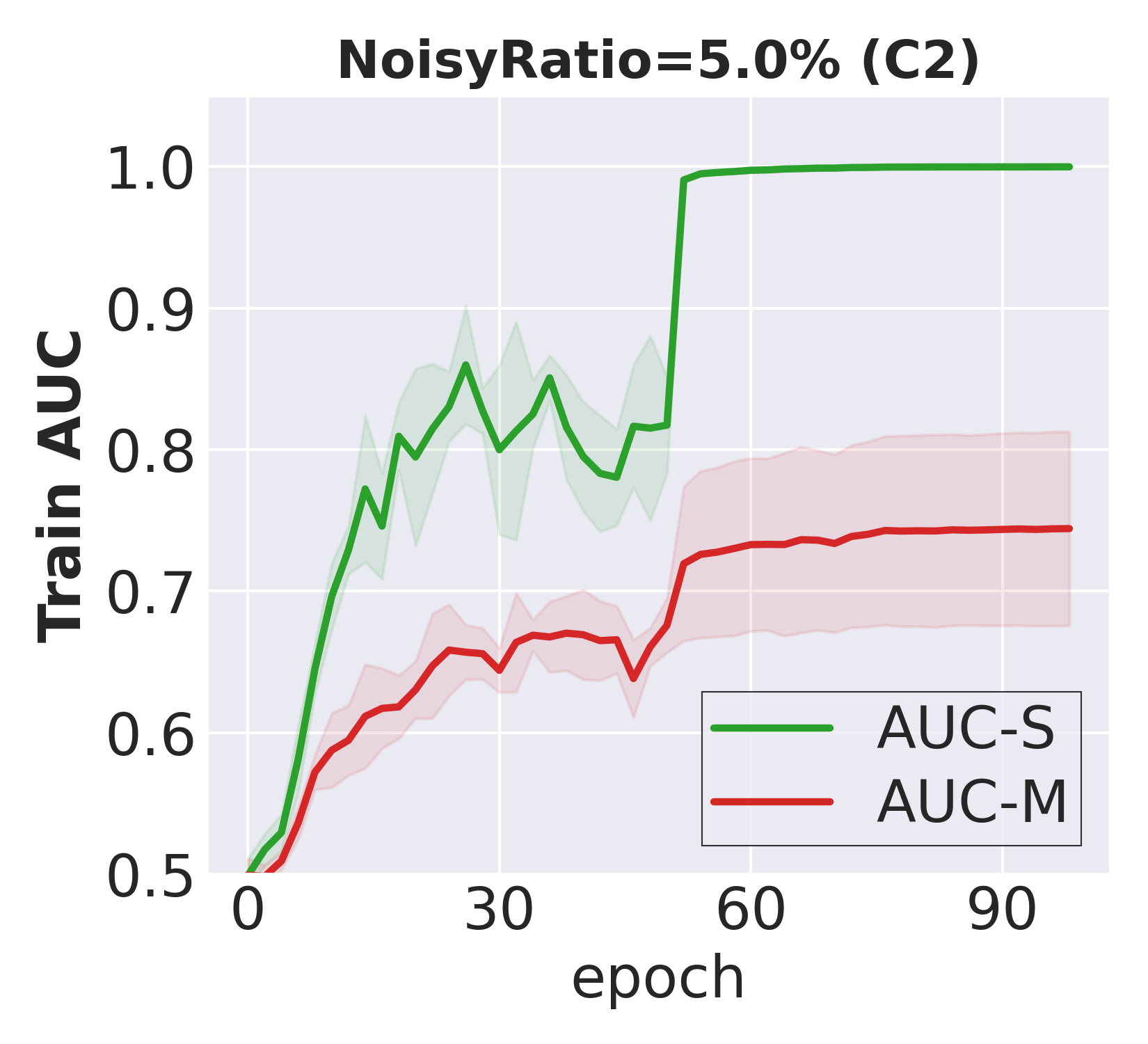}
\includegraphics[width=0.24\textwidth]{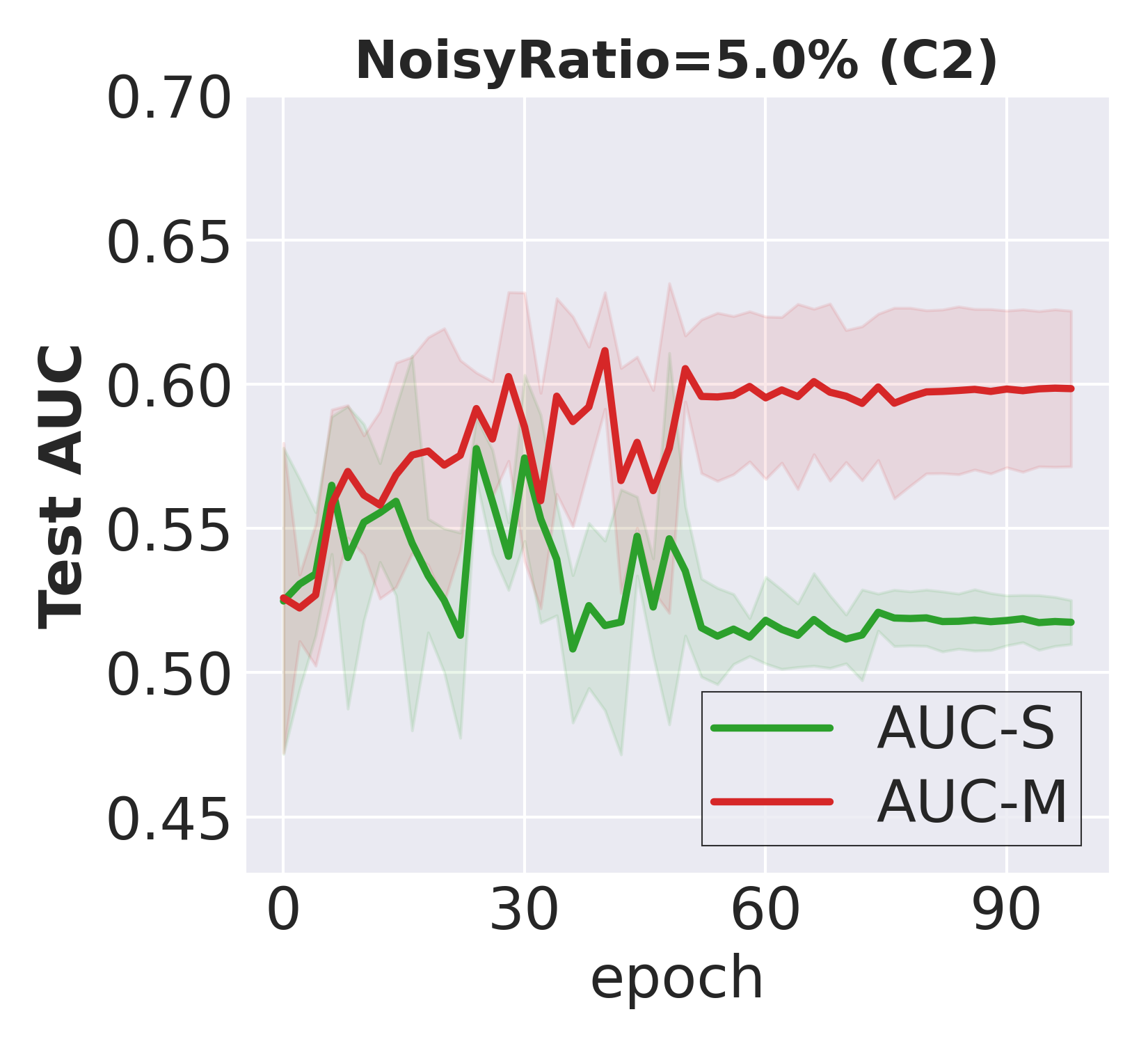}
\includegraphics[width=0.24\textwidth]{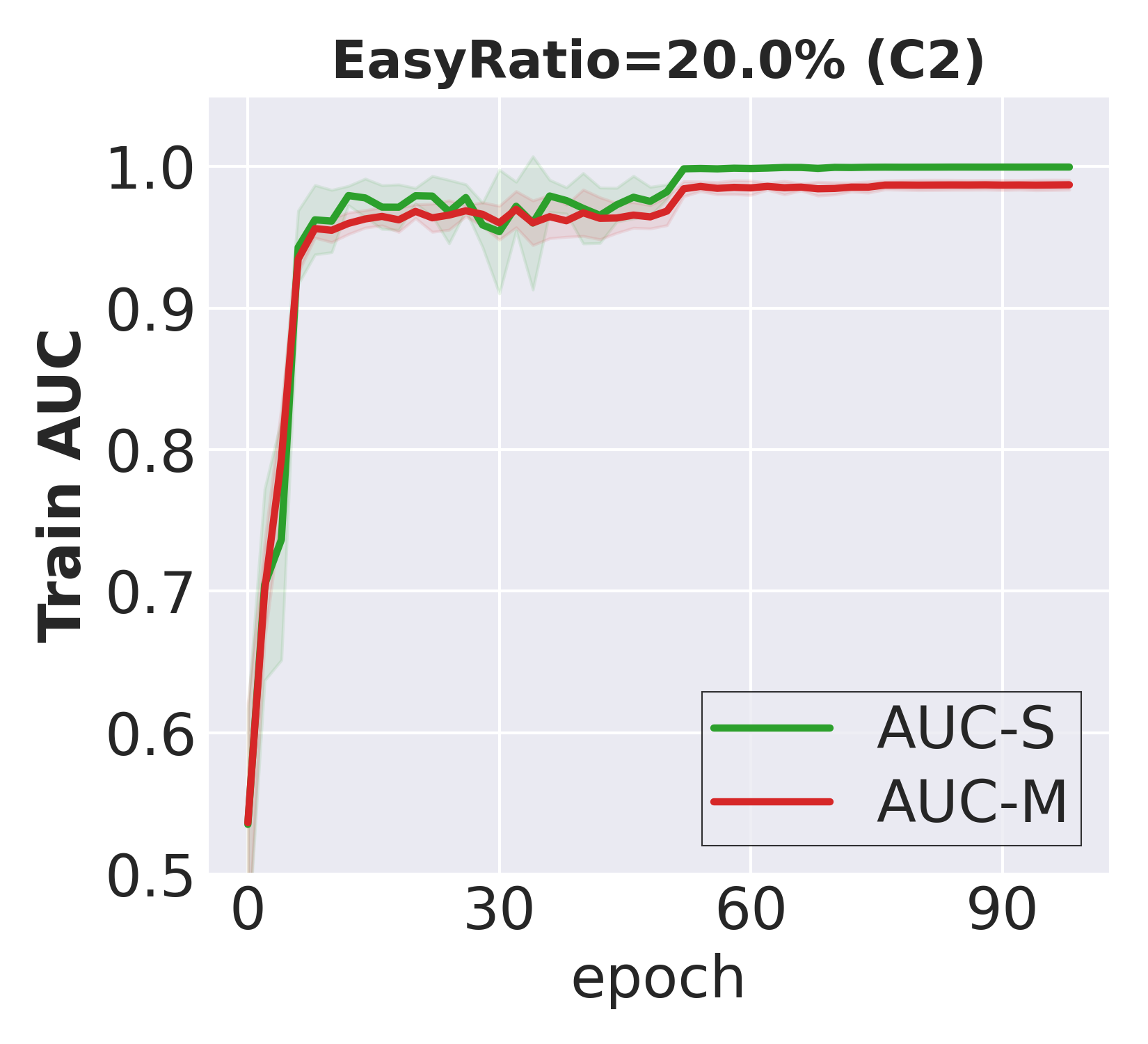}
\includegraphics[width=0.24\textwidth]{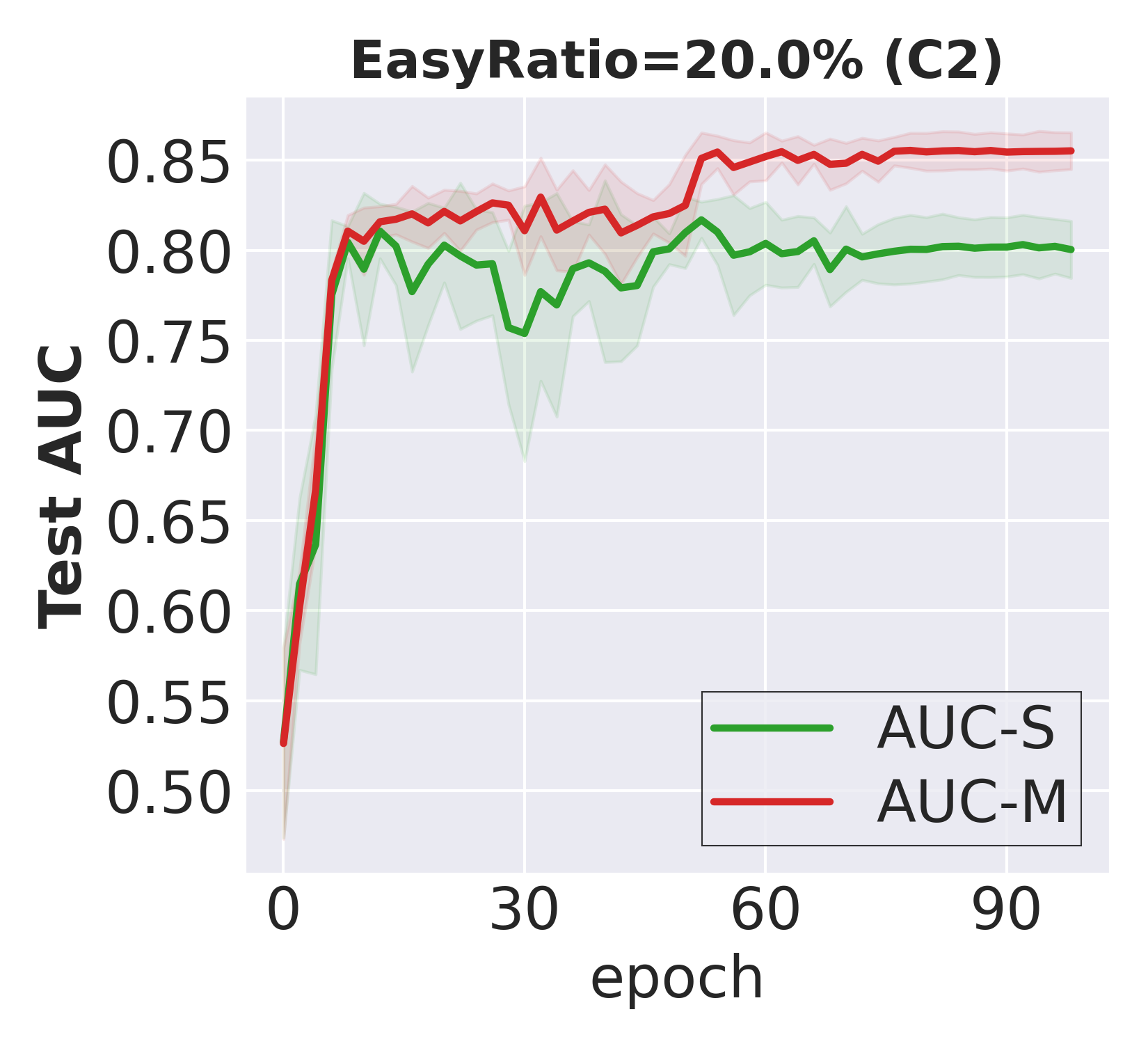}
\caption{Comparison when adding extra 20\% easy samples and 5\% noisy samples.}
\label{fig:noisy_easy_example_more}
\end{figure}

{\bf Robustness to Noisy Data and Easy Data.} We conduct ablation studies on the C2-IB data. To verify the robustness of our AUC-M loss to noisy data, we manually create some data with noisy labels. We construct the noisy dataset by modifying the C2 (imratio=1\%). To this end, we sample $1\%$ and $5\%$ from negative class to flip their labels to positive, and also randomly sample $1\%$ and $5\%$ positive data from the deleted positive examples and flip their labels and add them to the training data. This gives us two datasets with $1\%$ and $5\%$ noisy ratio. To verify the robustness of our AUC loss to easy data, we first pre-train a model by minimizing CE loss on C2 (imratio=1\%) and then  we make predictions on the removed positive samples and sort all prediction scores in descending order. Finally, we choose top 10\%, 20\% of sorted samples and add them to training data. We train DenseNet121 using batch size of 128 and initial learning rate of 0.1. Other parameter settings are the same as in Section~\ref{section:bechmark_results}.  We run experiments 5 times and plot the average testing AUC curve in Figure \ref{fig:noisy_easy_example} for the setting with $1\%$ noisy data and $10\%$ easy data. In Figure \ref{fig:noisy_easy_example_more}, we report results on other settings. All results clearly show that AUC-M outperforms AUC-S by a large margin.

{\bf Effect of Alpha Constraint.}
To verify the effectiveness of non-negative constraint on $\alpha$, we design an experiment to compare the performance of AUC-M with and without $\alpha \geq 0$ constraint. We start with C2-IB with imbalance ratio of 1\% and add 40\% easy (positive) samples and 1\% noisy samples to the training set similar to that is done above. 
We fix margin $m=0.1$. The curve of testing AUC and the curve of $\alpha$ v.s. \# of epochs are plotted in Figure \ref{fig:noisy_easy_example} (bottom panel). We observe that the performance with enforcing $\alpha \geq 0$ is better than that without enforcing it. The bottom right plot in Figure \ref{fig:noisy_easy_example} gives us a better illustration about the change of $\alpha$ during training. 
The plot inside it reveals the change of $\alpha$ in the first 2 epochs. It shows that the constraint prevents the value of $\alpha$ from dropping to a bad region and hence yields a faster convergence and better result. 	
	
\section{Conclusion}
In this paper, we have considered large-scale robust deep AUC maximization. We have proposed a new margin-based surrogate loss for AUC to address the two major issues of square loss, and demonstrated  its robustness to noisy and easy data. We thoroughly evaluate our methods on four benchmark datasets and four real-world medical datasets. The results not only demonstrate the effectiveness of the new margin loss and also the success of our deep AUC maximization methods on medical image classification tasks. 

\section*{Acknowledgements}
We are grateful to the anonymous reviewers for their constructive comments and suggestions. This work is partially supported by TY's NSF CAREER Award 1844403.

\bibliography{egbib}


\appendix
\newpage

\section{Optimal Values of $a, b,$ $\alpha$ in AUC Square Loss}
\label{section:optimal_a_b_alpha}
\vspace{-0.1in}
In Section \ref{section:drawbacks_of_AUC_square_loss}, we use the optimal values of $a, b, \alpha$.
In this section, we show how to derive these values.
We first re-present the min-max problem in (\ref{opt:spp}) as follows
\begin{align*}
\min_{\w\in\R^d \atop (a,b)\in\R^2}\max_{\alpha\in\R}f\left(\w,a,b,\alpha\right):=\EX_{\z}\left[F\left(\w,a,b,\alpha;\z\right)\right],
\end{align*}
where 
\begin{align*}	&F(\w,a,b,\alpha;\z)=(1-p)\left(h_\w(\x)-a\right)^2\mathbb{I}_{[y=1]}\\
&+p(p(1-p) + h_\w(\x)-b)^2\mathbb{I}_{[y=-1]}-p(1-p)\alpha^2\notag\\
&+2\alpha\left( p h_\w(\x)\mathbb{I}_{[y=-1]} - (1-p)h_\w(\x)\mathbb{I}_{[y=1]}\right) \notag .
\end{align*}
Given a fixed $\w$,the variable $a$ is only involved in the first term in $F$, so we have the $a$-subproblem as 
\begin{align*}
\min_a &
\E_\z [ (1-p) ( h_\w(\x) - a )^2 \I_{[y=1]} ]  
\\
= &
(1-p) \E_\z [ ( h_\w(\x) - a )^2 ] \cdot \E_z [ \I_{[y=1]} ]
\\
= &
(1-p) \E_\z [ ( h_\w(\x) - a )^2 | y=1 ] \cdot p  .
\end{align*}
As can be seen, $\E_\z [ ( h_\w(\x) - a )^2 | y=1 ]$ achieves minimum value when $a = \E[h_\w(\x) | y=1]$, which becomes the variance of $h_\w(\x)$. The optimal value of $b = \E[ h_\w(\x) | y = -1 ]$ can be achieved in the same way as $a$. The subproblem of $\alpha$ is 
\begin{align*}
\max_\alpha &
\E_\z [ 2\alpha( p(1-p) + p h_\w(\x) \I_{[y=-1]} - (1-p) h_\w(\x) \I_{[y=1]}) ]
- p(1-p) \alpha^2  
\\
= &
2\alpha ( p(1-p) + p \E_z [ h_\w(\x) \I_{[y=-1]} ] - (1-p) \E_\z [ h_\w(\x) ] \I_{[y=1]} )
- p(1-p) \alpha^2
\\
= &
2\alpha ( p(1-p) + p(1-p) \E_z [ h_\w(\x) | y=-1 ] - p(1-p) \E_\z [ h_\w(\x) | y=-1 ] )
- p(1-p) \alpha^2
\\
= &
p(1-p) \cdot ( 1 + 2\alpha ( \E_z [ h_\w(\x) | y=-1 ] - \E_\z [ h_\w(\x) | y=-1 ] ) 
- \alpha^2 )
\end{align*}
where we can derive its optimal value simply setting its gradient as zero.
This leads to
\begin{align*}
\alpha^*
= & 
1 + \E_z [ h_\w(\x) | y=-1 ] - \E_\z [ h_\w(\x) | y=-1 ]
\\
= &
1 + b(\w) - a(\w)   .
\end{align*}

\vspace{-0.15in}
\section{Reformulation of AUC Square Loss}
\label{section:reformulation_auc_square_loss}
\vspace{-0.1in}
In this section, we reformulate AUC square loss as follows
\begin{align}
A_\textS(\w) 
= &
\E[ ( 1 - h_\w(\x) + h_\w(\x') )^2 | y = 1, y' = -1 ]
\nonumber\\
= &
\E[ ( 1 - a(\w) + a(\w) - h(\w; \x) + h(\w; \x') - b(\w) + b(\w) )^2 | y = 1, y' = -1 ]
\nonumber\\
= &
\E[ \big( (a(\w)-h(\w; \x) + h(\w; \x') - b(\w)) + (1+b(\w)-a(\w))^2 \big) | x = 1, y' = -1 ]
\nonumber\\
= &
\E[ ( a(\w)-h(\w; \x) + h(\w; \x') - b(\w) )^2 
        + (1+b(\w)-a(\w))^2 
\nonumber \\
& ~~~~  + 2 \big( a(\w)-h(\w; \x) + h(\w; \x') - b(\w) \big) \cdot (1+b(\w)-a(\w))
       | y=1, y'=-1 ]
\nonumber\\
\stackrel{(e1)}{=} &
\E[ (h(\w; \x) - a(\w))^2 
        + (h(\w; \x') - b(\w))^2 
        - 2(h(\w; x) - a(\w)) \cdot (h(\w; \x') - b(\w) )
\nonumber\\
&  
        + (1+b(\w)-a(\w))^2 | y = 1, y' = -1 ]
\nonumber\\
\stackrel{(e2)}{=} &
\E[(h(\w; \x) - a(\w))^2 | y = 1]
+ \E[(h(\w; \x') - b(\w))^2 | y' = -1]
\nonumber\\
&
+ (1+b(\w)-a(\w))^2  
\nonumber\\
\stackrel{(e3)}{=} &
\E[(h(\w; \x) - a(\w))^2 | y = 1]
+ \E[(h(\w; \x') - b(\w))^2 | y' = -1]
\nonumber\\
&
+ \max_\alpha 2 \alpha (1+b(\w)-a(\w)) - \alpha^2   ,
\nonumber
\end{align}
where equality $(e1)$ is due to the definitions $a(\w) = \E[h(\w; \x) | y = 1]$ and $b(\w) = \E[h(\w; \x') | y' = -1]$, $\E[a(\w)] = a(\w)$ and $\E[b(\w)] = b(\w)$ ($a(\w)$ and $b(\w)$ are expectations, so they are constants).Equality $(e2)$ is due to the independence of the positive and negative samples. Equality $(e3)$ is due to the convex conjugate of the square function:
\begin{align*}
x^2 = \max_y 2 y \cdot x - y^2 .
\end{align*}

\vspace{-0.2in}
\section{Proof of Theorem \ref{thm:AUCM} }
\label{section:proof_of_auc_margin_equivalence}
Below, we start from the min-max problem and prove it is equivalent to the AUC margin loss in (\ref{eqn:AUCM}).
\begin{align}
\min_{a, b} \max_{\alpha \geq 0}
&
\E_\z [ F_{\text{M}} (\w, a, b, \alpha; \z) ]
\nonumber\\
= \min_{a, b} \max_{\alpha \geq 0}
&
\E_\z \Bigg[ (1-p) \left( h_\w(\x) - a \right)^2 \mathbb{I}_{[y=1]}
+ p ( h_\w(\x) - b )^2 \mathbb{I}_{[y=-1]} 
- p(1-p) \alpha^2
\nonumber\\
& + 2\alpha\left(p(1-p)m+ p h_\w(\x)\mathbb{I}_{[y=-1]}-(1-p)h_\w(\x)\mathbb{I}_{[y=1]}\right) \Bigg]
\nonumber\\
= \min_{a, b} \max_{\alpha \geq 0}
&
\Bigg[
(1-p) \E_\z [ \left( h_\w(\x) - a \right)^2 \mathbb{I}_{[y=1]} ]
+ p \E_\z [ ( h_\w(\x) - b )^2 \mathbb{I}_{[y=-1]} ]
- p(1-p) \alpha^2
\nonumber\\
& 
+ 2\alpha\left(p(1-p)m 
+ p \E_\z [ h_\w(\x) \mathbb{I}_{[y=-1]} ]
- (1-p) \E_\z [ h_\w(\x) \mathbb{I}_{[y=1]} ] \right) 
\Bigg]
\nonumber\\
= \max_{\alpha \geq 0} \qquad
&
p(1-p) \Bigg[ \E_\z [ \left( h_\w(\x) - a(\w) \right)^2 | y = 1 ]
+ \E_\z [ ( h_\w(\x) - b(\w) )^2 | y = -1 ]
- \alpha^2 
\nonumber\\
\label{eq:recover_auc_margin_loss}
& 
+ 2\alpha\left( m + b(\w) - a(\w) \right) \Bigg]
=
p(1-p) A_\textM(\w)
\\
= \qquad \qquad
&
p(1-p) \Bigg[ \E_\z [ \left( h_\w(\x) - a(\w) \right)^2 | y = 1 ]
+ \E_\z [ ( h_\w(\x) - b(\w) )^2 | y = -1 ]
+ \left( m + b(\w) - a(\w) \right)^2_+ \Bigg]
\nonumber
\nonumber
\end{align}
where (\ref{eq:recover_auc_margin_loss}) shows the equivalence between minimizing $A_{\text{M}}(\w)$ in (\ref{eqn:AUCM}) and 
$\underset{\w,a,b}{\min} \underset{\alpha \geq 0}{\max} \E_\z [ F_{\text{M}} (\w, a, b, \alpha; \z) ]$, i.e.,
\begin{align}
\min_{\w, a, b} \max_{\alpha \geq 0} \E_\z [ F_\textM(\w, a, b, \alpha; \z) ] = p(1-p)
A_\textM(\w).
\nonumber
\end{align}
The last equality is to explicitly show the squared hinge loss.

\section{Analysis of Adverse Effect on Easy Data of Square loss based on the min-max formulation}\label{app:easyminmax}
In particular, the gradient  of $F(\w, a, b, \alpha;\z)$ is given by 
$\nabla_\w F(\w, a, b, \alpha; \z) =2(1-p)\x \I_{[y=1]} \cdot (h_\w(\x) - a - \alpha)+ 2p\x \I_{[y=-1]} \cdot (h_\w(\x) - b + \alpha) $.
When $\z$ is positive, the first term above is active, by plugging the optimal value of $a, b, \alpha$ given $\w$, the stochastic gradient descent update will yields an updated model as 
\begin{align*}
\w_+ = \w - \eta 2(1-p)\x \I_{[y=1]} \cdot (h_\w(\x) - 1 -  b), 
\end{align*}
where $b$ is the mean prediction score on negative  data.  When $\x$ is an easy positive data such that $h_\w(\x) - 1 -  b>0$, then $\w_+$ will move towards the negative direction of the positive data $\x$, as a result it will  push the score $h_{\w_+}(\x)$ on the positive data smaller than $h_\w(\x)$, which is harmful for AUC maximization. Similarly, we have the same phenomenon when the sampled data $\z$ is negative. 

\section{A $1$-Dim Example of Easy/Noisy Data for AUC Square and Margin Loss}
\label{section:example_easy_noisy_data}
Suppose we have a $1$-dimensional AUC maximization problem with a linear model parameterized by a $1$-dimensional model $\w$, i.e., $h_\w(\x) = \w \cdot \x$, so that $\nabla_\w h_\w(\x) = \x$.
Recall the definition of $F$ in (\ref{opt:spp}), we have its gradient w.r.t. $\w$ as follows
\begin{align*}
\nabla_\w F(\w, a, b, \alpha; \z)
= &
2(1-p) ( h_\w(\x) - b ) \cdot \nabla_\w h_\w(\x) \I_{[y=1]}
+ 2p ( h_\w(\x) - b ) \cdot \nabla_\w h_\w(\x) \I_{[y=-1]}
\\
&
+ 2\alpha ( p\nabla_\w h_\w(\x) \I_{[y=-1]} - (1-p) \nabla_\w h_\w(\x) \I_{[y=1]} )
\\
= &
2(1-p) \nabla_\w h_\w(\x) \I_{[y=1]} \cdot \underbrace{ ( h_\w(\x) - a - \alpha ) }_{ = B }
+ 2p \nabla_\w h_\w(\x) \I_{[y=-1]} \cdot \underbrace{ ( h_\w(\x) - b + \alpha ) }_{ = C }  ,
\end{align*}
where our study focuses on the two terms $B$ and $C$, which determines the direction of $\nabla_\w F$ for $y=1$ and $y=-1$, respectively.

$\alpha$ is the key difference between AUC square loss in (\ref{opt:spp}) and AUC margin loss in (\ref{eqn:AUCM}).
To simplify the explanation, we let $a = a(\w)$ and $b = b(\w)$ achieve their optimal values.
In AUC square loss (\ref{opt:spp}), $\alpha$ is not constrained, and the optimal value is $\alpha = 1 + b - a$.
In AUC margin loss (\ref{eqn:AUCM}), it has a non-negative constraint on $\alpha$, so the optimal value is $\alpha = \max\{ 0, 1 + b - a \}$.

\subsection{Easy Data for AUC Square Loss}
\label{subsection:easy_auc_s}

At the $t$-th iteration, let $\w_t = 1$ and we have two easy data $(\x_1=1, y_1=1)$ and $(\x_2=-1, y_2=-1)$.
We assume that $a = 0.5$ and $b = -0.5$.

For $(\x_1, y_1 = 1)$
\begin{align*}
B 
= 
h_\w(\x_1) - a - \alpha
=
h_\w(\x_1) - 1 - b
=
1 \times 1 - 1 - (-0.5)
=
0.5  ,
\end{align*}
which indicates that $\nabla_\w F \propto \nabla_\w h_\w(\x_1)$ (they are in the same direction).
By assuming all the constants and the step size can be merged into a constant value $0.1$, the stochastic gradient descent can be 
\begin{align*}
\w_{t+1} 
=
\w_t - 0.1 \times \nabla_\w h_{\w_t} (\x_1)
=
1 - 0.1 \times \x_1
=
1 - 0.1 \times 1
=
0.9.
\end{align*}
Then we re-evaluate the prediction score by $\w_{t+1}$:
\begin{align*}
h_{\w_{t+1}}(\x_1)
=
0.9 \times 1
=
0.9 
< 
h_{\w_t} (\x_1)
= 1  .
\end{align*}
In this case, the prediction score for a positive sample decreases, which is an undesirable update.

For $(\x_2, y_2 = -1)$
\begin{align*}
C 
= 
h_\w(\x_1) - b + \alpha
= 
h_\w(\x_1) + 1 - a
=
-1 + 1 - 0.5
=
-0.5  ,
\end{align*}
which indicates that $\nabla_\w F \propto -\nabla_\w h_\w(\x_1)$ (they are in the negative direction of each other).
By assuming all the constants and the step size can be merged into a constant value $0.1$, the stochastic gradient descent can be 
\begin{align*}
\w_{t+1} 
=
\w_t - 0.1 \times (-1) \times \nabla_\w h_{\w_t} (\x_1)
=
1 + 0.1 \times \x_2
=
1 + 0.1 \times (-1)
=
0.9.
\end{align*}
Then we re-evaluate the prediction score by $\w_{t+1}$:
\begin{align*}
h_{\w_{t+1}}(\x_2)
=
0.9 \times (-1)
=
-0.9 
> 
h_{\w_t} (\x_2)
= -1  .
\end{align*}
In this case, the prediction score for a negative sample increases, which is an undesirable update.

\subsection{Easy Data for AUC Margin Loss}
\label{subsection:easy_auc_m}

Since the optimal $\alpha = \max\{ 0, m + b - a \}$, we consider the two cases, respectively.

{\bf Case 1: $\alpha = 0$.}
This case indicates that $m + b - a \leq 0$ or $m + b \leq a$, which is a good situation, because $a$ (the mean prediction of positive data) and $b$ (the mean prediction of negative data) are sufficiently far away from each other by a margin of $m$.
Here for simplicity, we assume that at the $t$-th iteration, $\w = 1$, $m=1$, $a = 1$ and $b=-0.5$.

For $(\x_1 = 0.75, y=1)$:
\begin{align*}
B 
=
h_\w(\x_1) - a - \alpha
=
h_\w(\x_1) - a
=
0.75 - 1
=
-0.25   \qquad \text{(negative direction)} ,
\end{align*}
where $h_\w(\x_1) > m + b = 0.5$ means that $\x_1$ is well classified, but $F_\textM$ still suffers a penalty on it and push it to be closer to $a=1$.

For $(\x_1 = 1.25, y=1)$:
\begin{align*}
B 
=
h_\w(\x_1) - a - \alpha
=
h_\w(\x_1) - a
=
1.25 - 1
=
0.25   \qquad \text{(negative direction)} ,
\end{align*}
where $h_\w(\x_1) > m + b = 0.5$ means that $\x_1$ is well classified, but $F_\textM$ still suffers a penalty on it and push it to be closer to $a=1$.
To sum up, when the model is good enough, i.g., $m + b < a$, $F_\textM$ only push positive data towards $a$ and negative data towards $b$.

{\bf Case 2: $\alpha = m + b - a$.}
This case indicates that $m + b - a > 0$ or $m + b > a$, which is a undesirable situation, because $a$ (the mean prediction of positive data) and $b$ (the mean prediction of negative data) are within a margin of $m$.
Here for simplicity, we assume that at the $t$-th iteration, $\w=1, m=1, a=0, b=-0.5$.

For $(\x_1 = 0.25, y_1 = 1)$:
\begin{align*}
B
=
h_\w(\x_1) - a - \alpha
=
h_\w(\x_1) - m - b
=
0.25 - 1 + 0.5
=
-0.25   \qquad \text{(negative direction)} ,
\end{align*}
where $h_\w(\x_1) < m + b = 0.5$ means that $\x_1$ is not well classified.
Thus, the stochastic gradient descent for updating $\w_t$ can be
\begin{align*}
\w_{t+1} 
= 
\w_t - 0.1 \times (-1) \times \nabla_w h_{\w_t}(\x_1)
=
1 + 0.1 \times \x_1
= 
1 + 0.1 \times 0.25
=
1.025   ,
\end{align*}
which makes the prediction of $\x_1$ larger: $h_{\w_{t+1}}(\x_1) = 1.025 \times 0.25 = 0.2562 > h_{\w_t}(\x_1) = 0.25$.

Examples for negative data can be derived in a similar way, so we omit those presentation.

\subsection{Noisy Data for AUC Square Loss}
\label{subsection:noisy_auc_s}

Assuming $\w=1$, consider the case where $m + b > a$, i.e., the model is not good, e.g., $a =0.25, b=-0.5$.
For $(\x_1=0.25, y_1=-1, y_1^\true=1)$, since only $y_1$ is revealed, we will use term $C$ to determine $\nabla_\w F$.
On the other hand, since $y_1^\true=1$, we know that $h_\w(\x)$ can be large.
Then we can compute its term $C$
\begin{align*}
C
=
h_\w(\x_1) - b + \alpha
=
h_\w(\x_1) + 1 - a
=
0.25 \times 1 + 1 - 0.25
=
1   \qquad \text{(positive direction)} ,
\end{align*}
which means that $\nabla_\w F$ is in the same direction of $\nabla_w h_\w(\x_1)$.
It is exactly the same case in Section \ref{subsection:easy_auc_s} when $B > 0$, so it will give an undesirable update.

Negative sample $(\x_2=-1, y_1=1, y_1^\true=-1)$ can be developed in the same way, which also gives an undesirable update.

\subsection{Noisy Data for AUC Margin Loss}
\label{subsection:noisy_auc_m}

Assuming $\w = 1$, consider the case where $m + b > a$, i.e., the model is not good, and $\alpha = m + b - a$.
We assume $a = 0.25, b=-0.5$.
For $(\x_1=0.25, y_1=-1, y_1^\true=1)$:
\begin{align*}
C 
=
h_\w(\x_1) -b + \alpha
=
h_\w(\x_1) -b + ( m + b - a )
=
h_\w(\x_1) + m - a
=
0.25 \times 1 + m - 0.25
=
m   .
\end{align*}
$m$ is positive by dentition.
However, unlike the previous AUC square loss where $m=1$, in AUC margin loss $m$ is a hyper-parameter.
Even though we cannot completely resolve the noisy data issue by using AUC margin loss, we can still reduce the magnitude of update along with the wrong direction by changing $m$ to a smaller value from constant $1$.

The same situation happens for noisy negative data on the not-so-good model.

\section{An Example of Sensitivity of AUC}
\label{section:example_sensitivity_auc_accuracy}
\begin{table}[h]
\centering
\caption{Illustrations of sensitivity of Accuracy and AUC on an imbalanced dataset of 25 samples with a positive ratio of 3/25. The accuracy threshold is 0.5. \textbf{Example 1} shows that all positive instances rank higher than negative instances and two negative instances are misclassified to positive class. \textbf{Example 2} shows that 1 positive instance ranks lower than 7 negative instances and 1 positive and 1 negative instances are missclassifed.  \textbf{Example 3} shows that 2 positive instances rank lower than 7 negative instances, and 2 positive instances are also missclassifed as negative class. Overall, we can observe that AUC drops dramatically as the ranks of positive instances drop but meanwhile Accuracy remains unchanged.}
\vspace{0.05in}
\scalebox{0.9}{
\begin{tabular}{cccccccc}
\cline{1-4} \cline{4-7} \cline{7-8}
\multicolumn{2}{c}{\textbf{Example 1}} &           & \multicolumn{2}{c}{\textbf{Example 2}} &           & \multicolumn{2}{c}{\textbf{Example 3}} \\ \cline{1-4} \cline{4-7} \cline{7-8} 
\text{Prediction}    & \text{Ground Truth}   & \text{} & \text{Prediction}    & \text{Ground Truth}   & \text{} & \text{Prediction}    & \text{Ground Truth}   \\ \cline{1-4} \cline{4-7} \cline{7-8} 
0.9 & 1 &  & 0.9 & 1 &  & 0.9 & 1 \\ 
0.8 & 1 &  & {\bf 0.41}($\downarrow$)  & 1 &  & {\bf 0.41}($\downarrow$) & 1 \\  
0.7 & 1 &  & 0.7 & 1 &  & {\bf 0.40}($\downarrow$) & 1 \\ 
0.6 & 0 &  & 0.6 & 0 &  & {\bf 0.49}($\downarrow$) & 0  \\
0.6 & 0 &  & {\bf 0.49}($\downarrow$) & 0 &  & {\bf 0.48}($\downarrow$)& 0 \\ 
0.47 & 0 &  & 0.47 & 0 &  & 0.47 & 0 \\ 
0.47 & 0 &  & 0.47 & 0 &  & 0.47 & 0 \\
0.45 & 0 &  & 0.45 & 0 &  & 0.45 & 0 \\
0.43 & 0 &  & 0.43 & 0 &  & 0.43 & 0 \\
0.42 & 0 &  & 0.42 & 0 & & 0.42 & 0 \\ 
\vdots & \vdots  &  & \vdots  & \vdots  & & \vdots  & \vdots  \\ 
0.1 & 0 &  & 0.1 & 0 & & 0.1 & 0 \\ 

\cline{1-4} \cline{4-7} \cline{7-8} 
\multicolumn{2}{c}{\begin{tabular}[c]{@{}c@{}}Acc=0.92  \\ AUC=1.00 \end{tabular}} & \multicolumn{1}{l}{} & \multicolumn{2}{c}{\begin{tabular}[c]{@{}l@{}}Acc=0.92 (---) \\  AUC={\bf 0.89} ($\downarrow$) \end{tabular}} & \multicolumn{1}{c}{} & \multicolumn{2}{c}{\begin{tabular}[c]{@{}c@{}}Acc=0.92 (---)   \\AUC={\bf 0.78} ($\downarrow$)  \end{tabular}} \\\cline{1-4} \cline{4-7} \cline{7-8} 
\end{tabular}}
\label{tab:auc_sensitive}
\end{table}

\newpage
\section{Descriptions of Imbalanced Datasets}
\label{section:dataset_description}
\vspace{-0.2in}
\begin{table}[h]
\centering
\caption{Description of of Datasets. Note that ''size of training set" refers to the number of samples for the original training set. Datasets with suffix ''-IB" denote that we manually construct the imbalanced datasets by randomly removing some positive samples.}
\scalebox{0.8}{
\begin{tabular}{ccc}
\hline
\textbf{Datasets}     & \textbf{Size of image} & \textbf{Size of training set}                        \\ \hline
Cat\&Dog-IB & low resolution &  25,000  \\ 
CIFAR10-IB  &  low resolution  & 50,000 \\ 
CIFAR100-IB &  low resolution &  50,000\\ 
STL10-IB    & medium resolution & 5,000 \\ 
PatchCamelyon-IB & medium resolution  & 294,912 \\
Melanoma    & high resolution &  46,131  \\ 
CheXpert    & high resolution  & 223,416 \\
DDSM+ & high resolution  & 55,890 \\ \hline
\end{tabular}
\label{tab:datasets_summary}}
\end{table}

\section{More Experiments on Benchmark Datasets}
\label{section:more_benchmar_results}
\begin{figure}[h]
\centering
\includegraphics[width=0.22\textwidth]{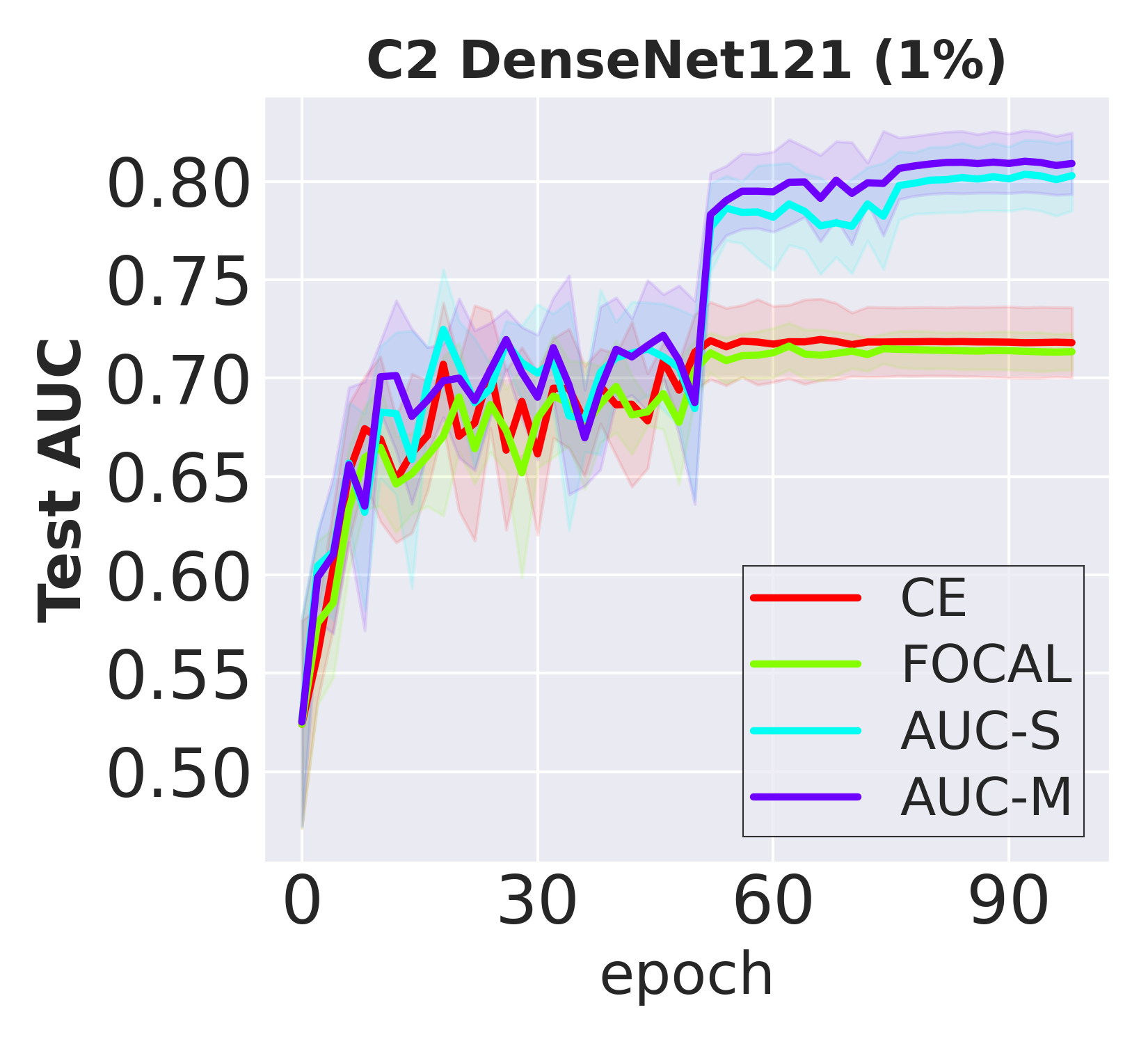}
\includegraphics[width=0.22\textwidth]{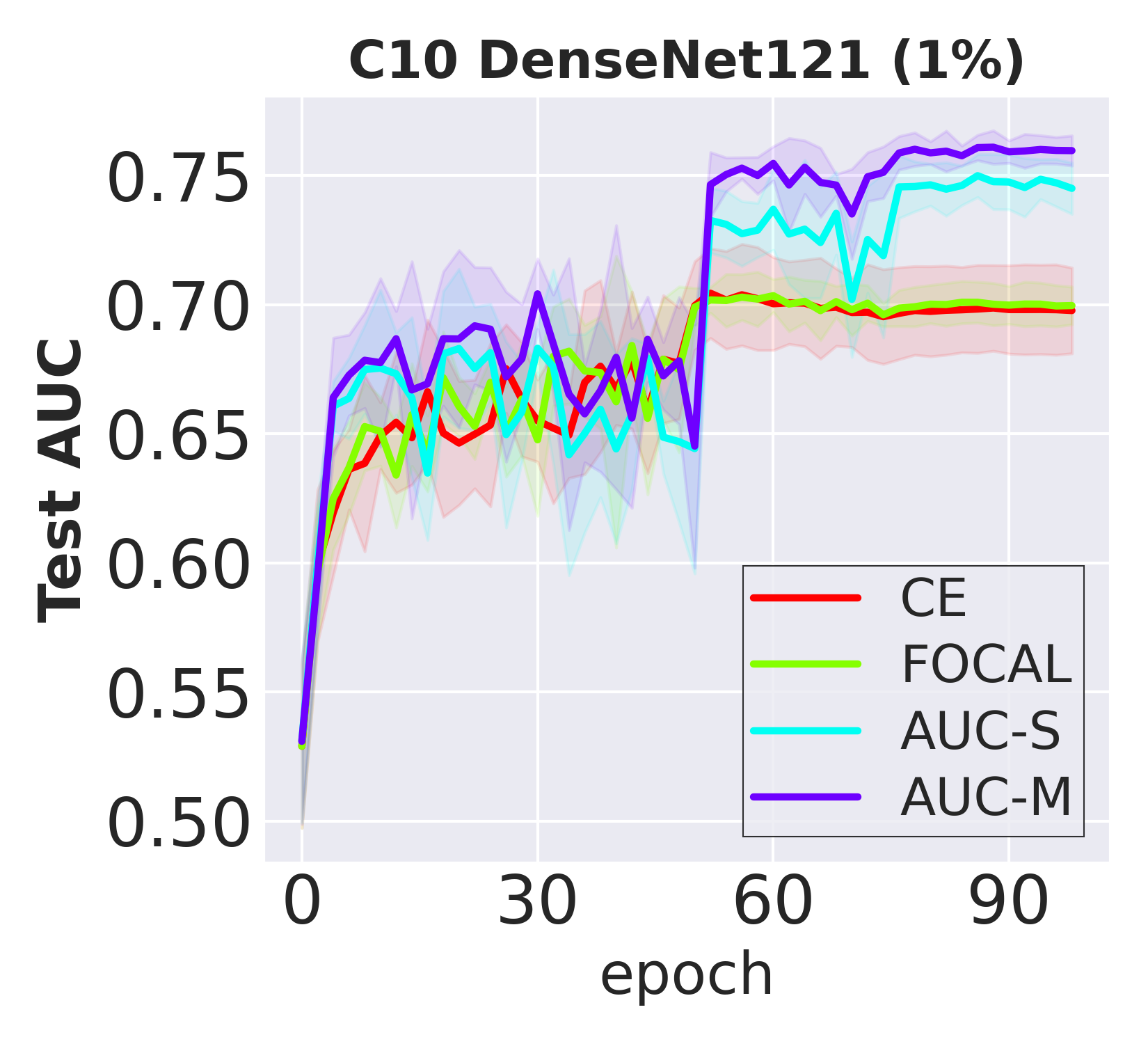}
\includegraphics[width=0.22\textwidth]{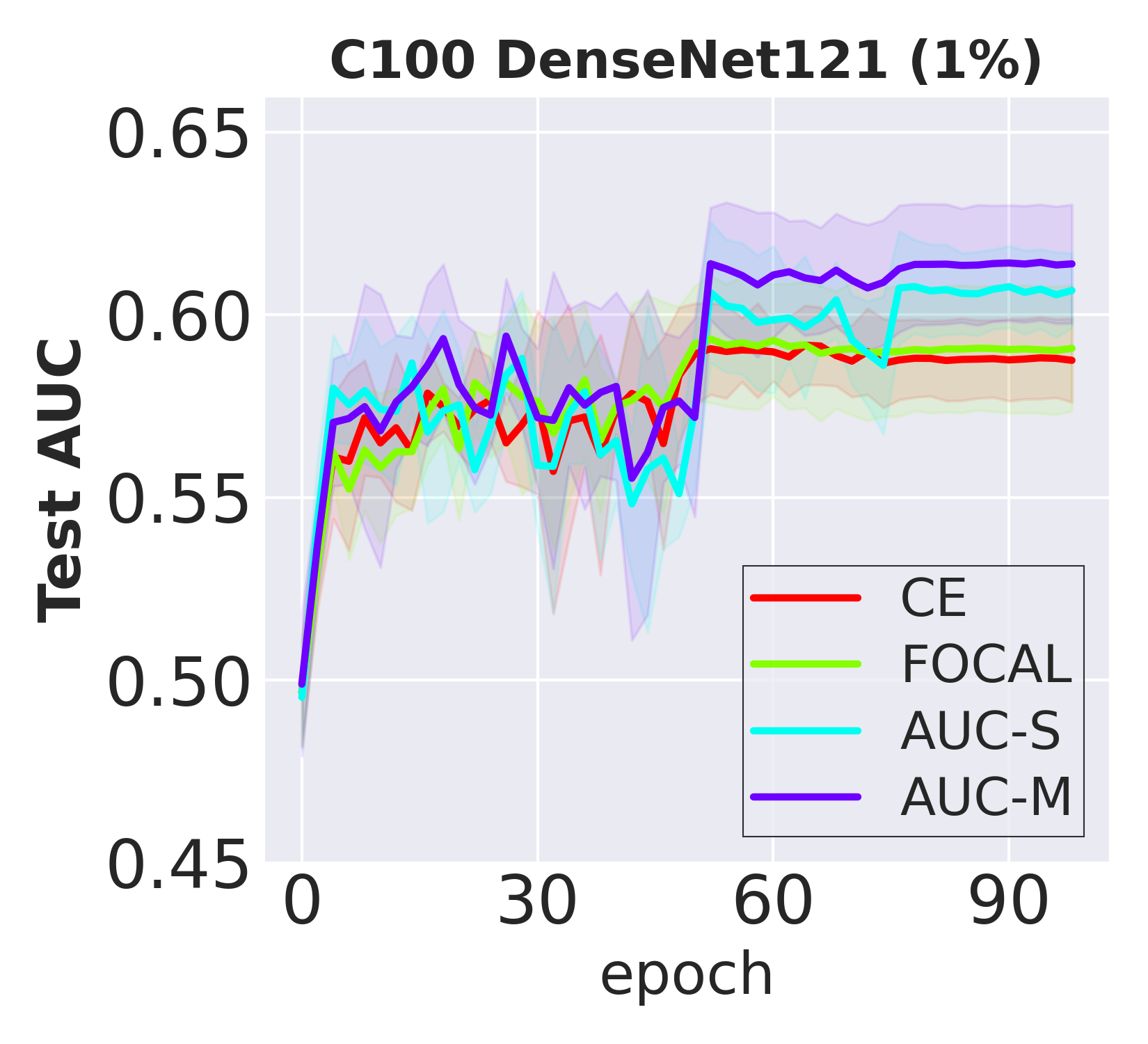}
\includegraphics[width=0.22\textwidth]{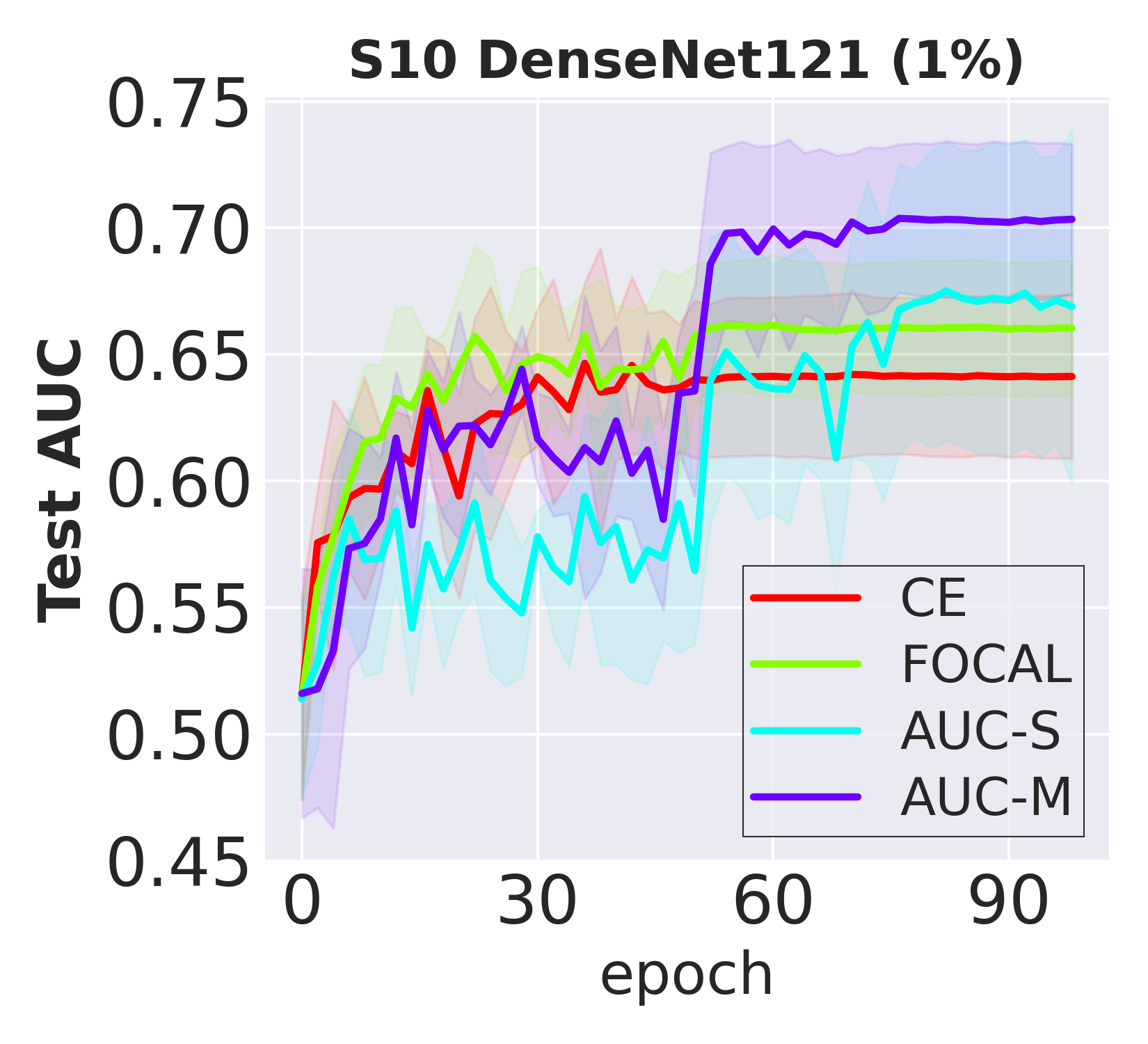}
\caption{Testing AUC vs epochs on Benchmark Datasets for DenseNet121. }
\label{fig:bechmark_auc}
\end{figure}

\begin{figure}[h]
\centering
\includegraphics[width=0.22\textwidth]{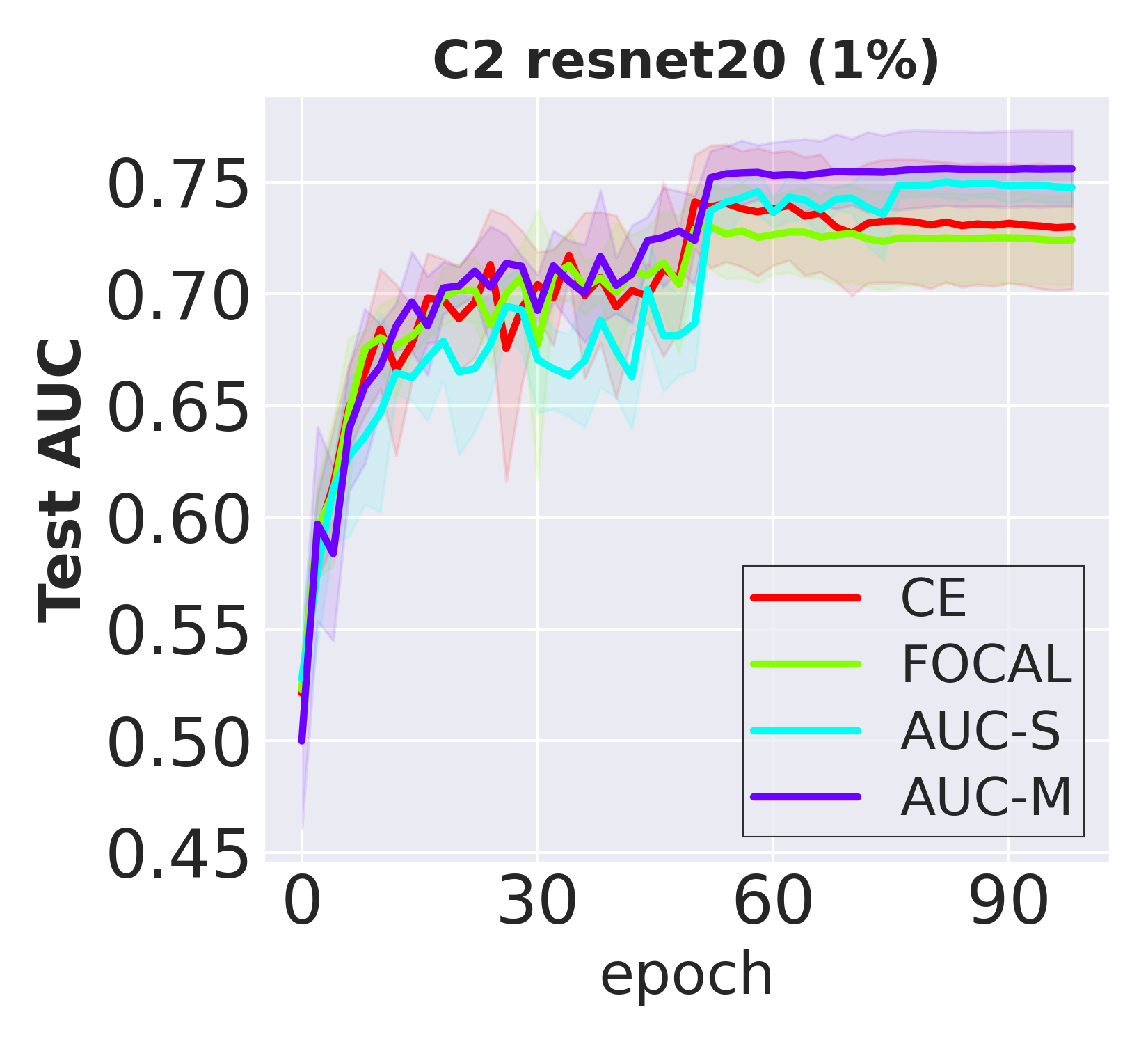}
\includegraphics[width=0.22\textwidth]{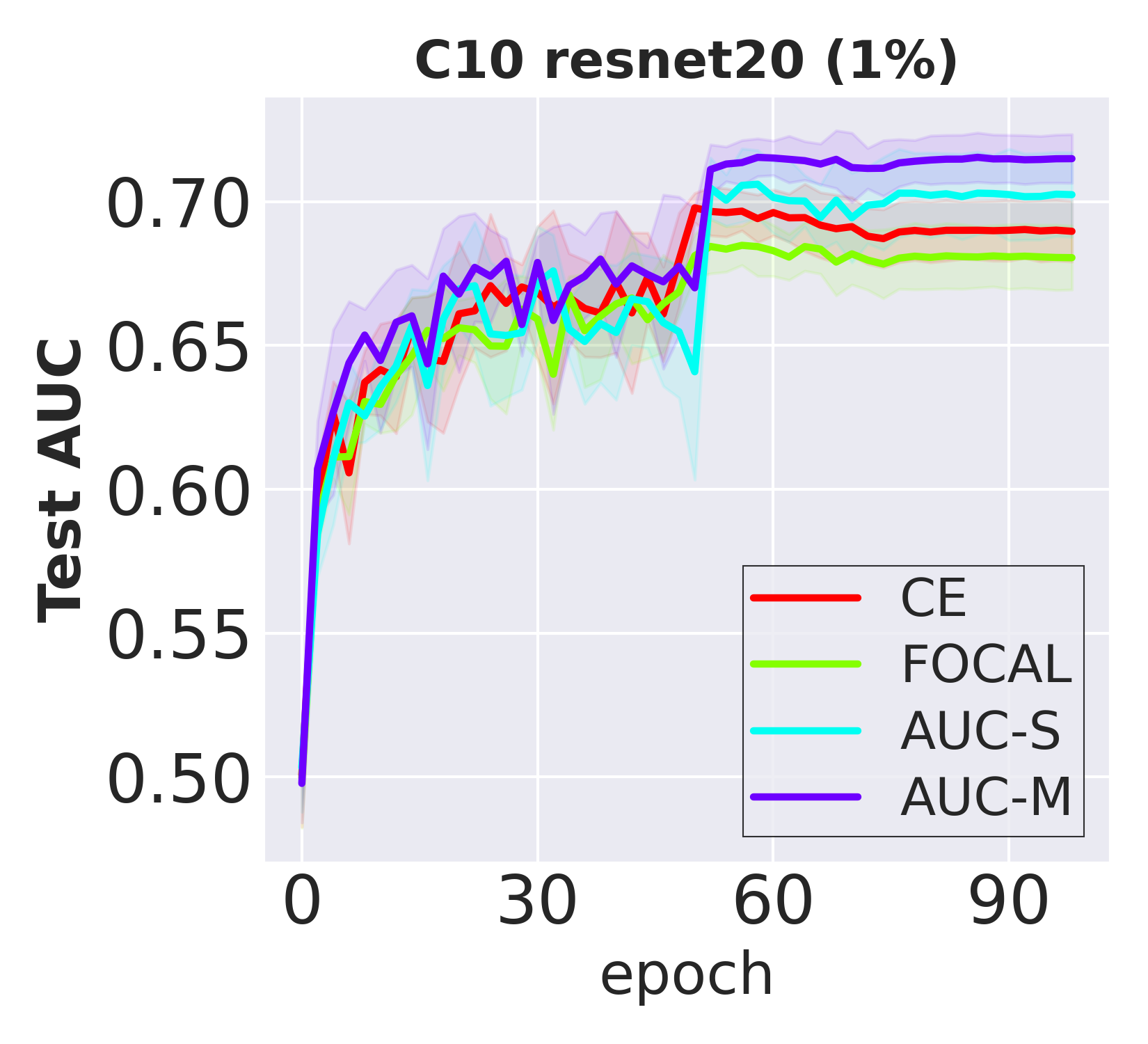}
\includegraphics[width=0.22\textwidth]{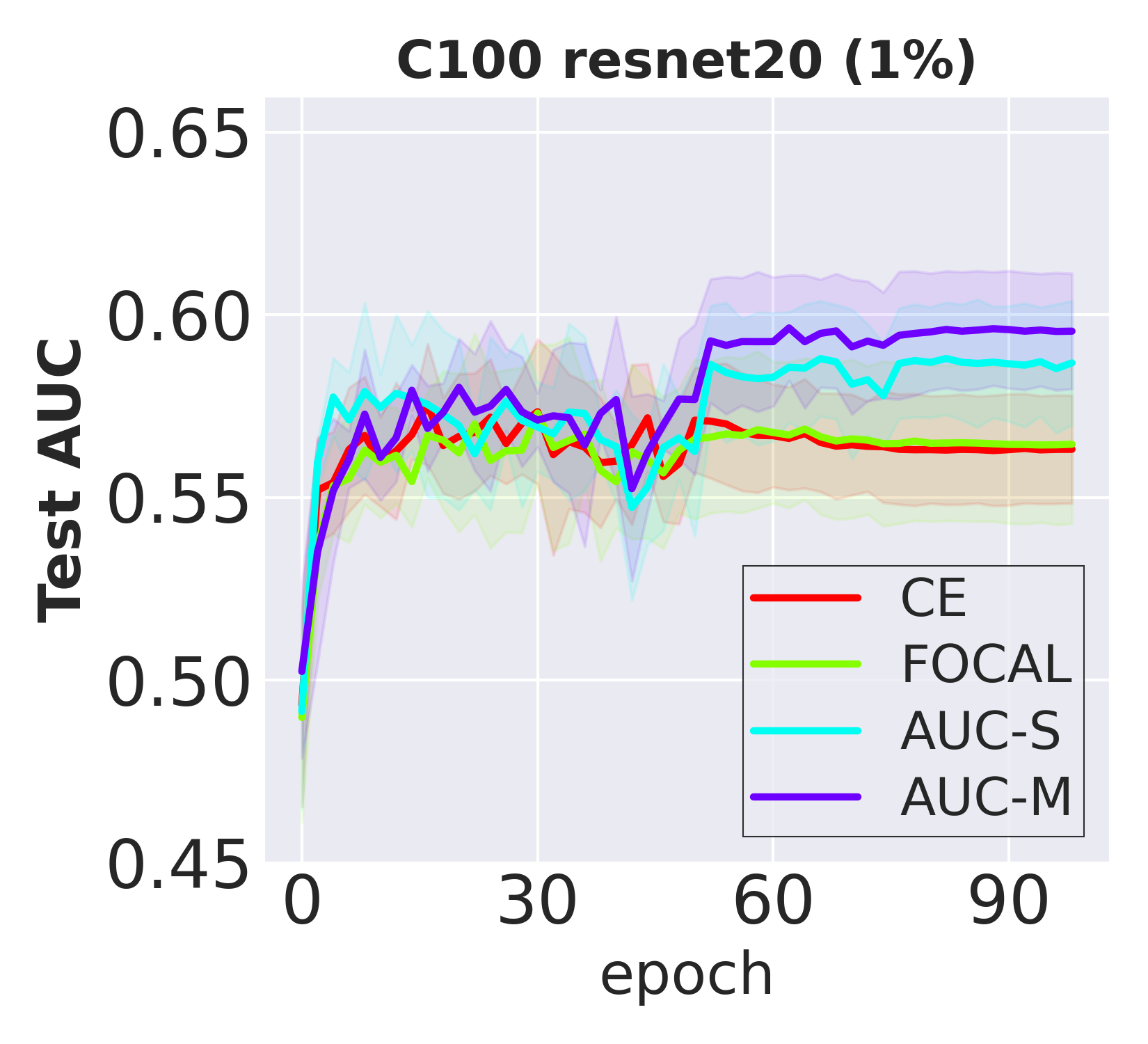}
\includegraphics[width=0.22\textwidth]{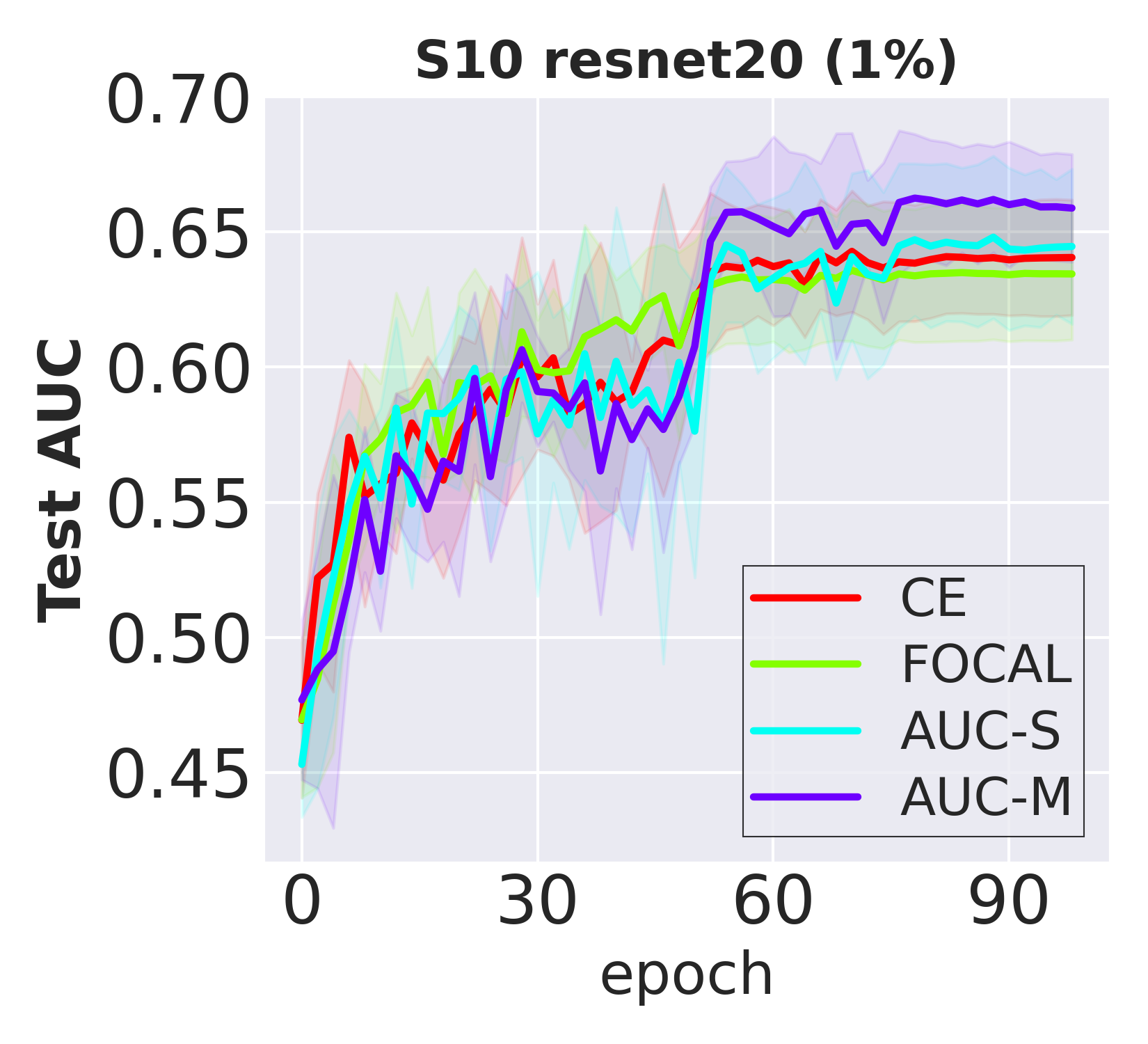}
\caption{Testing AUC vs epochs on Benchmark Datasets for ResNet20.}
\label{fig:bechmark_auc_resnet}
\end{figure}

\begin{table}[h]
\centering
\caption{Testing AUC of benchmark datasets with DenseNet121(D) and ResNet20(R) for imratio=10\%. Note that when the imbalance ratio increases e.g., from $1\%$ to $10\%$, data becomes less imbalanced and the classification becomes easier. }
\label{tab:results_benchmark_resnet}
\scalebox{0.85}{
\begin{tabular}{ccccccc}
\hline
\textbf{Dataset} & \textbf{imratio} & \textbf{CE} & \textbf{Focal} & \textbf{AUC-S} & \textbf{AUC-M}  \\ \hline
C2 (D)       &  10\%          & 0.893$\pm$0.004          & 0.879$\pm$0.005    & 0.901$\pm$0.002 & \textbf{0.902$\pm$0.001}   \\ 
C10 (D)      & 10\%           & \textbf{0.898$\pm$0.005} & 0.879$\pm$0.005    & 0.889$\pm$0.002          & 0.887$\pm$0.005 \\ 
S10 (D)     & 10\%            & 0.820$\pm$0.015          & 0.819$\pm$0.010    & 0.825$\pm$0.013          & \textbf{0.846$\pm$0.015}  \\  
C100 (D)    & 10\%           & 0.710$\pm$0.007          & 0.705$\pm$0.007    & 0.720$\pm$0.003 & \textbf{0.723$\pm$0.006}      \\ \hline
C2 (R)       & 10\%     & 0.920$\pm$0.004 & 0.881$\pm$0.008 & 0.897$\pm$0.007& {\bf0.920$\pm$0.006}\\
C10 (R)      & 10\%     & {\bf 0.898}$\pm$0.004 & 0.851$\pm$0.018 & 0.872$\pm$0.007 & 0.898$\pm$0.005 \\
S10 (R)     & 10\%     & {\bf 0.825}$\pm$0.013 & 0.813$\pm$0.009 & 0.819$\pm$0.013 & 0.821$\pm$0.011 \\
C100(R)      & 10\%     &  0.669$\pm$0.006 & 0.666$\pm$0.012 &  0.686$\pm$0.005 & {\bf0.695$\pm$0.003} \\ \hline
\end{tabular}}
\end{table}

\newpage
\section{The Choice of Margin $m$ for AUC-M Loss}
Margin $m$ is an important parameter for AUC-M loss. As illustrated in Section 3.3, when the model is not good enough, noisy data may produce a stochastic gradient that indicates a wrong direction. In this case, a smaller $m$ can alleviate such sensitivity to noisy data. Tuning $m$  parameter can trade off the margin benefit and the robustness to noisy data. That is the reason why tuning $m$ is important in AUC-M. On benchmark datasets, the average values of $m$ over different random trials are  0.7,0.8,0.7,0.5  on C2, C10, S10, C100, respectively. On Melanoma, the best $m$ is 0.8. On CheXpert, the best $m$ is 0.8 in average over 5 classes. On DDSM, the best $m$ is 0.5. On PatchCamelyon, the best $m$ is 0.7. For the results of ablation studies, we use $m=0.3$ for AUC-M loss.

\section{A Two-stage Training Framework for DAM}
\label{section:two_stage_auc_max_framework}
\begin{figure*}[h]
\begin{center}
\includegraphics[width=0.7\textwidth]{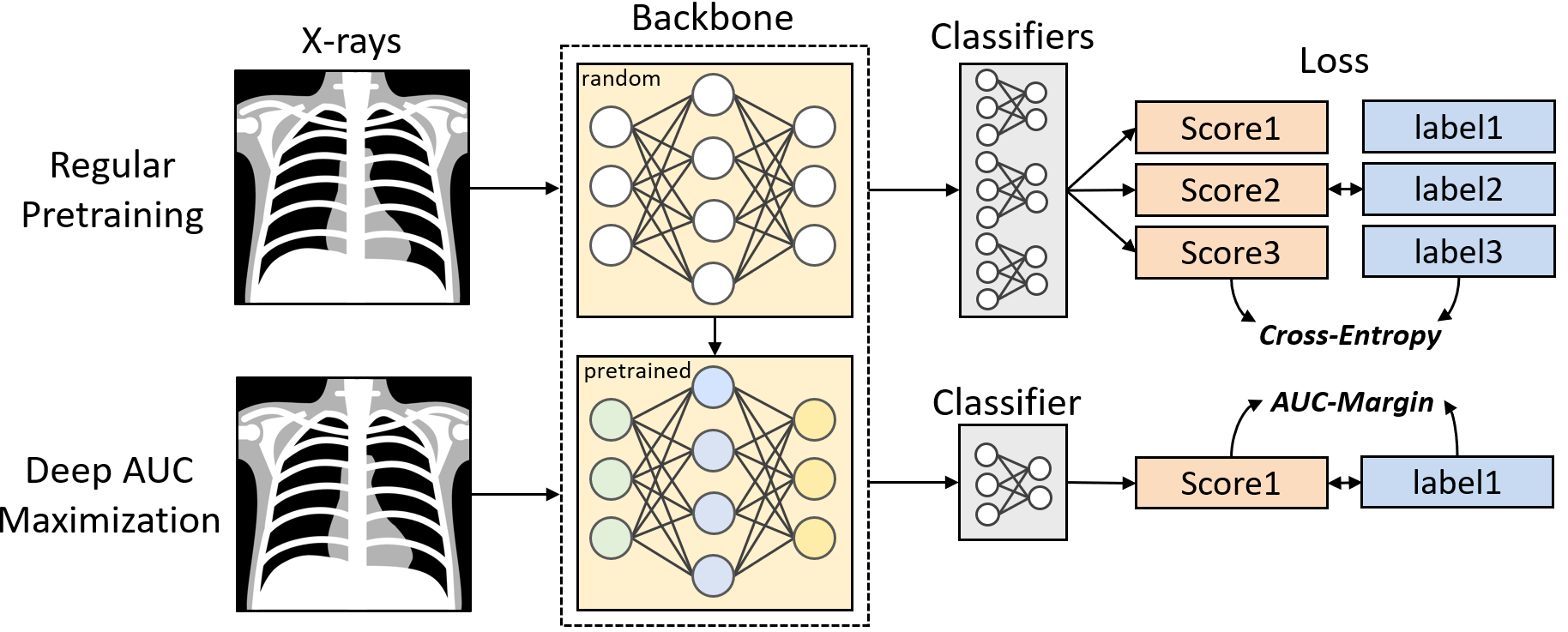}
\end{center}
\caption{A Two-stage Deep AUC Maximization Framework. For the pretraining stage, we focus on learning representation by optimizing a standard CrossEntropy loss. For the AUC maximization stage, we focus on finetuning the decision boundary of classifier by optimizing AUC margin loss. }
\label{fig:two_stage_training_frameowork}
\end{figure*}

\section{Network Architecture for Melanoma Classification}
\label{section:more_results_on_kaggle}
\begin{figure}[h]
\begin{center}
\includegraphics[width=0.5\textwidth]{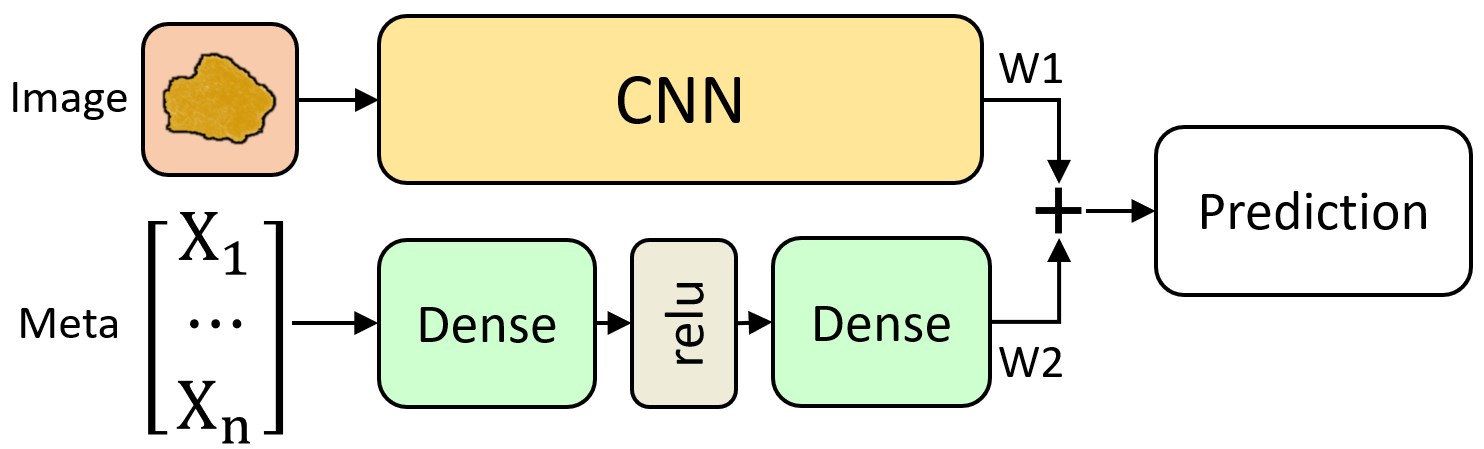}
\end{center}
\caption{A mixed network architecture of a CNN (EfficientNet) and a 2-layer Neural Network for predicting Melanoma using image and patient contextual data. For training, we first train the CNN model and then train DNN model (using same configurations) but freeze the parameter updates for CNN model. The training configurations are described in main section.}
\label{fig:kaggle_training_frameowork}
\end{figure}

\end{document}